\documentclass{article}

\usepackage{arxiv}

\usepackage[utf8]{inputenc} 
\usepackage[T1]{fontenc}    
\usepackage{hyperref}       
\usepackage{url}            
\usepackage{booktabs}       
\usepackage{amsfonts}       
\usepackage{nicefrac}       
\usepackage{microtype}      
\usepackage{lipsum}		
\usepackage{graphicx}
\usepackage{natbib}
\usepackage{doi}
\usepackage{todonotes}
\usepackage{amsmath}
\usepackage{subcaption}
\usepackage{float}
\usepackage{adjustbox}
\usepackage{booktabs,tabularx,siunitx,amsmath,amssymb}
\title{Generative Urban Flow Modeling:\\ From Geometry to Airflow with Graph Diffusion}

\date{} 					

\author{
\textbf{Francisco Giral}$^{1}$\thanks{Corresponding author: fa.giral@alumnos.upm.es} \quad \textbf{Álvaro Manzano}$^{1,2}$ \quad \textbf{Ignacio Gómez}$^1$ \\
\textbf{Petros Koumoutsakos}$^3$ \quad \textbf{Soledad Le Clainche}$^1$ \\
\small $^1$ Applied Mathematics Department, ETSIAE-School of Aeronautics, Universidad Politécnica de Madrid, Spain \\
\small $^2$ Microflown Technologies, Tivolilaan 205, 6824BV Arnhem, Netherlands \\
\small $^3$ Computational Science and Engineering Laboratory, Harvard University, 29 Oxford Street, Cambridge, MA 02138, USA \\
}

\renewcommand{\undertitle}{}
\renewcommand{\shorttitle}{Generative Urban Flow Modeling: From Geometry to Airflow with Graph Diffusion}
\hypersetup{
pdftitle={Generative Urban Flow Modeling: From Geometry to Airflow with Graph Diffusion},
pdfsubject={},
pdfauthor={},
pdfkeywords={},
}

\begin{document}
\maketitle

\begin{abstract}
Urban wind flow modeling and simulation play an important role in air quality assessment and sustainable city planning. A key challenge for modeling and simulation is handling the complex geometries of the urban landscape. Low order models are limited in capturing the  effects of geometry, while high-fidelity Computational Fluid Dynamics (CFD) simulations are prohibitively expensive, especially across multiple geometries or wind conditions. Here, we propose a generative diffusion framework for synthesizing steady-state urban wind fields over unstructured meshes that requires only geometry information. The framework combines a hierarchical graph neural network with score-based diffusion modeling to generate accurate and diverse velocity fields without requiring temporal rollouts or dense measurements. Trained across multiple mesh slices and wind angles, the model generalizes to unseen geometries, recovers key flow structures such as wakes and recirculation zones, and offers uncertainty-aware predictions. Ablation studies confirm robustness to mesh variation and performance under different inference regimes. This work develops is the first step towards foundation models for the built environment that can  help urban planners rapidly evaluate design decisions under densification and climate uncertainty.
\end{abstract}


\section{Introduction}
\label{sec:introduction}

Urban wind flows are critical for air quality, pedestrian comfort, heat stress, and pollutant dispersion within cities. They are affected by the arrangement of buildings and street canyons that produce complex, multiscale flow structures \citep{blocken2015computational, world2016ambient,Nature2025}. City planners often resort to modeling to estimate wind patterns in urban environments. The fidelity of these simulations  is  essential for climate-responsive urban planning, real-time air-quality assessment, and the planning of healthier,innovative and resilient built environments.
Computational Fluid Dynamics (CFD) solvers that rely on Reynolds-Averaged Navier--Stokes (RANS) and Large Eddy Simulation (LES) can reproduce key flow features including recirculation zones, shear layers, and vortex shedding \citep{lateb2016use}. However, these solvers require substantial computational effort: even a single steady RANS simulation on a city-scale mesh may take hours to days on high-performance computing (HPC) systems, while LES at comparable resolution can require days to weeks of wall-clock time. This makes them unsuitable for iterative urban-design workflows, scenario exploration, or operational digital-twin systems. Surrogates and low order models are an  alternative that aim to produce reusable representations of flow behavior to  amortize the cost of simulation across multiple geometries and inflow conditions.

Data-driven surrogates based on simplified machine-learning models have been applied to approximate urban canopy wind distributions \citep{BenMoshe2023}. Deep generative models such as generative adversarial networks (GANs) have enabled real-time synthesis of 2D flow fields \citep{Kastner2023GAN}, while physics-informed graph neural networks (GNNs) have incorporated RANS residuals into message passing over unstructured city meshes to improve generalization \citep{Shao2023PIGNN}. Multi-scale surrogate frameworks have learned to downscale coarse meteorological fields to neighborhood-scale flow patterns \citep{Lu2023CanopyFlows}. Alongside these developments, probabilistic diffusion models have demonstrated strong performance in fluid reconstruction and uncertainty-aware surrogate modeling \citep{Yang2023FluidDiff, Shu2023DiffusionPDE, Whittaker2024TurbulenceDiffusion, Vishwasrao2025DiffSPORT}. These advances mirror broader trends in the application of generative models to weather forecasting \citep{Price2024GenCast, bonev2025fourcastnet, bodnar2025foundation} and parameterized turbulent simulation \citep{Gao2025CMA_GuidedDiffusion}.
Despite significant progress, existing approaches for urban-wind surrogate modeling share two fundamental limitations. First, most learning-based surrogates require case-specific flow information at inference time. They are typically trained as spatiotemporal predictors that are conditioned on one or more reference flow snapshots (e.g., models that take $(X_{t-1}, X_{t})$ as input and predict $X_{t+1}$, where $X$ denotes the fluid-field variables to be generated); on partial sensor observations; or on coefficients in a reduced modal basis \citep{Yang2023FluidDiff, Lu2023CanopyFlows, Vishwasrao2024diffusion, Gao2025CMA_GuidedDiffusion, barragan2025hybrinethybridneuralnetworkbasedframework}. In contrast, the present method does not rely on any CFD-derived flow information at inference. Instead, it generates a physically plausible converged steady-state field for a given urban configuration using only the input mesh and global parameters such as inflow direction.
Second, many existing methods rely on structured grids or fixed spatial layouts, which limits their applicability to the highly irregular meshes used in CFD practice. One line of work mitigates this by projecting the solution onto a lower-dimensional latent space, for example, through modal decomposition \citep{abadia2022predictive} or learned neural embeddings \citep{Wang2024CoNFiLD}. Although such techniques can, in principle, be formulated on unstructured meshes, in practice they often require interpolating the CFD solution to a reference grid or to a global modal basis. This decoupling can blur sharp obstacle boundaries and street-canyon interfaces, introduces additional modeling choices regarding the mesh-to-latent mapping, and may inject spurious interpolation noise. In contrast, GNN-based learned simulators have shown a strong capacity to handle unstructured spatial domains and model fluid–obstacle interactions in spatiotemporal settings \citep{valencia2025learning, pfaff2021mgn, fortunato2022msmgn, Brandstetter2022MP-PDE, Nguyen2025PhysiX, Gao2025GraphLED}. Operating directly on the unstructured CFD mesh preserves obstacle-aware resolution and avoids interpolation artifacts. This is crucial when the surrogate is intended to replace, rather than be periodically corrected by, expensive CFD evaluations.

To the best of our knowledge, there is currently no surrogate modeling approach that can generate steady-state wind fields on new unstructured meshes, where building presence and configuration vary. Existing solutions require case-specific flow snapshots, sensor fields, or modal coefficients at inference time. Here, we propose a novel, generative learning  framework that models the probabilistic relationship between urban geometry and the resulting flow field. Notably this framework  enables the computation of steady-state wind fields directly from mesh geometry and inflow direction. The model operates on arbitrary unstructured meshes and is conditioned only on node coordinates, obstacle information, mesh connectivity, and inflow direction. The core of the method is a diffusion-based generative model whose denoising dynamics is parameterized by a hierarchical multiscale GNN \citep{pfaff2021mgn, fortunato2022msmgn}. This architecture learns a continuous probability distribution over physically plausible flow fields defined in spatial domains with varying topology. Because the model is trained across multiple altitude slices of a real city, where building presence and geometry vary with height, it learns to associate obstacle configuration with characteristic flow patterns. As a result, it generalizes to unseen mesh geometries and obstacle arrangements, allowing fast generation of wind fields in near real-time.

This work is a step in the development of foundation models for the built environment. As cities continue to densify and climate variability intensifies, urban planners and policymakers require tools that allow rapid evaluation of design decisions under uncertainty. The ability to generate physically plausible flow fields directly from geometry, without case-specific solver runs, opens the door to real-time microclimate assessment, interactive design interfaces, and continuous digital-twin monitoring. Moreover, because our formulation conditions on mesh topology rather than on specific spatial discretizations, it can naturally extend to multi-altitude, city-scale, or fully three-dimensional atmospheric domains, as well as to additional physical processes such as thermal exchange or pollutant dispersion. Coupled with automatically generated urban geometries from geographic information systems (GIS) or building-information models, such surrogates could support rapid what-if analyses for pollution episodes, adaptive traffic regulation, and emergency response in dense city centers.

Methodologically, our approach is motivated by the complementary strengths of diffusion models and multiscale graph neural networks. Diffusion models provide a principled probabilistic framework that learns an entire distribution of physically plausible flows, rather than a single deterministic mapping. This property is crucial for the present task: generating a steady-state wind field on a new urban configuration without any reference flow snapshot. Deterministic supervised models typically fail in this setting, as they require conditioning on previous flow states or partial observations to produce stable predictions. Diffusion models, in contrast, enable sampling directly from the learned geometry-conditioned distribution of steady solutions. However, existing diffusion-based fluid models have been applied only to structured grids or simplified domains, and, to our knowledge, have not yet been deployed on unstructured CFD meshes with the complex and variable topology characteristic of urban environments. Multiscale GNNs naturally complement diffusion processes by capturing interactions across spatial scales while operating directly on irregular meshes, preserving obstacle-aware detail without interpolation. By combining these components, the present method yields a surrogate that is both probabilistic and mesh-native, capable of producing realistic urban wind fields directly from geometry and global parameters alone.

In summary, this work introduces the first geometry-conditioned generative model for urban wind flows that operates on CFD-grade unstructured meshes and does not require flow snapshots or modal coefficients at inference time.
The method is demonstrated on a realistic high-Reynolds urban CFD dataset of a Bristol neighborhood, comprising multiple building configurations and 360 inflow angles and  generalizes well to unseen geometries. The framework is a foundation for developing fast, probabilistic CFD surrogates in urban applications, enabling real-time microclimate analyses, rapid what-if studies, and online decision support for air quality management and urban planning.

This paper is organized as follows. Section~\ref{sec:method} describes the methodology, including the model architecture, training procedure, and dataset. Section~\ref{sec:results} presents the experimental results. Section~\ref{sec:conclusion} summarizes the findings and outlines future research directions.

\section{Probabilistic Generation of Urban Flows}
\label{sec:method}
We first outline the physical characteristics of urban wind flows and the challenges inherent in their simulation and modeling. Wind flow in urban environments can be modelled using  the incompressible Navier–Stokes equations with complex boundary conditions due to sharply varying building geometries, terrain undulations, and anisotropic forcing from large-scale meteorology. Flow features such as separation, recirculation bubbles, corner vortices, channeling effects along street canyons, and wake interactions across multiple scales make the dynamics highly nonlinear and spatially heterogeneous. Resolving these structures typically requires advanced CFD approaches such as LES or even Direct Numerical Simulation (DNS), although the latter is rarely feasible in urban settings due to its extreme computational cost. In practice, steady-state RANS models are commonly used: while they provide only mean flow fields, these are sufficient to capture the key patterns of interest in urban aerodynamics, including recirculation zones, shear layers, corner vortices, and wake topology.

However, high-fidelity CFD in realistic urban domains is computationally prohibitive: a single RANS solution on a 40-million-cell mesh may require hours to days on HPC infrastructure, even for steady-state configurations. Exploring wind behavior over 360 different angles, as is often required in wind comfort analysis, dispersion studies, or risk assessment, can therefore demand extended time of computation and significant energy consumption. This cost severely limits scenario exploration, parametric studies, and rapid urban planning feedback loops.

These challenges motivate learning-based generative models capable of sampling physically plausible flow fields conditioned on external forcing, such as wind direction, without resorting to repeated CFD runs. However, urban flows are not deterministic mappings of external conditions: even steady RANS solutions exhibit inherent uncertainty due to turbulence modeling approximations and sensitivity to initialization. Thus, a probabilistic approach is more suitable than deterministic regression. Furthermore, the spatial domain is defined on irregular 2D slices derived from 3D geometries, making conventional convolution-based solutions unsuitable due to their reliance on uniform grid structures. These constraints necessitate a generative model that is (i) probabilistic, (ii) geometry-aware, and (iii) capable of operating on graph-based unstructured meshes.

\subsection{The Elucidated Diffusion Model (EDM)}
\label{sec:edm_background}

Diffusion models provide a probabilistic framework for learning complex data distributions through iterative denoising. They simulate a forward process that gradually perturbs samples from the data distribution $p_{\text{data}}(\mathbf{x})$ with Gaussian noise, and learn the inverse process that reconstructs clean samples from noise. A diffusion model defines a sequence of latent variables $\{\mathbf{x}_t\}_{t=0}^T$ such that
\begin{equation}
q(\mathbf{x}_t | \mathbf{x}_{t-1}) = 
\mathcal{N}\!\left(
\sqrt{1-\beta_t}\, \mathbf{x}_{t-1}, \, \beta_t \mathbf{I}
\right),
\label{eq:ddpm_forward}
\end{equation}
where $\beta_t \in (0,1)$ denotes the noise variance schedule. After the $T$ steps, $\mathbf{x}_T$ approaches an isotropic Gaussian distribution. The model then learns the reverse process, parameterized by a neural network, which progressively removes noise from $\mathbf{x}_T$ to recover a sample from $p_{\text{data}}$.

Although this formulation underpins the Denoising Diffusion Probabilistic Model (DDPM)~\cite{ho2020denoisingdiffusionprobabilisticmodels}, subsequent work revealed that many of its components, such as noise schedules, loss weights and scaling factors, can be unified and generalized. The Elucidated Diffusion Model (EDM)~\cite{karras2022edm} provides such a unified framework by decomposing the design space of diffusion models into modular and interpretable components.

EDM considers the data distribution $p_{\text{data}}(\mathbf{x})$ as part of a continuum of noisy distributions $p(\mathbf{x}; \sigma)$ obtained by perturbing data with Gaussian noise of standard deviation $\sigma$:
\begin{equation}
p(\mathbf{x}; \sigma) = 
\int p_{\text{data}}(\mathbf{y}) \,
\mathcal{N}\!\left(\mathbf{x} \mid \mathbf{y}, \sigma^2 \mathbf{I}\right)
\, d\mathbf{y}.
\end{equation}
As $\sigma$ increases from zero to a large value $\sigma_{\max}$, $p(\mathbf{x}; \sigma)$ seamlessly transitions from the data manifold to a standard Gaussian. The generative process can then be described as a deterministic probability flow ODE:
\begin{equation}
\frac{d\mathbf{x}}{dt} = 
- \frac{\dot{\sigma}(t)}{\sigma(t)} 
\nabla_{\mathbf{x}} \log p(\mathbf{x}; \sigma(t)),
\label{eq:edm_ode}
\end{equation}
where $\nabla_{\mathbf{x}} \log p(\mathbf{x}; \sigma)$ is the score function. Integrating this ODE backward from the maximum standard deviation ($\sigma_{\max}$) to the minimum standard deviation ($\sigma_{\min}) \approx 0$ yields samples distributed according to $p_{\text{data}}$.

The score function is approximated using a neural network $D_\theta(\mathbf{x}, \sigma)$ trained to predict the sample $\mathbf{x}_0$ given a noisy input $\mathbf{x} = \mathbf{x}_0 + \sigma \boldsymbol{\epsilon}$ with $\boldsymbol{\epsilon} \sim \mathcal{N}(0, \mathbf{I})$. The optimal denoiser minimizes
\begin{equation}
\mathcal{L}_{\text{EDM}} =
\mathbb{E}_{\mathbf{x}_0, \boldsymbol{\epsilon}, \sigma}
\!\left[
\lambda(\sigma)
\big\|
D_\theta(\mathbf{x}_0 + \sigma \boldsymbol{\epsilon}, \sigma)
- \mathbf{x}_0
\big\|_2^2
\right],
\label{eq:edm_loss}
\end{equation}
where $\lambda(\sigma)$ is a noise-dependent weighting term that stabilizes training across noise levels:
\begin{equation}
\lambda(\sigma) = 
\frac{(\sigma^2 + \sigma_{\text{data}}^2)}{(\sigma \, \sigma_{\text{data}})^2}.
\label{eq:edm_lambda}
\end{equation}
This weighting ensures that gradients are balanced between small and large noise levels, preventing instability or overfitting to low-noise examples. In Equation (\ref{eq:edm_lambda}) $\sigma$ represents the noise level, and $\sigma_{\text{data}}$ represents the standard deviation of the targets after normalization.

An important contribution of EDM is the preconditioning of the neural network’s inputs and outputs. Instead of directly predicting $\mathbf{x}_0$, the network is designed to output a scaled correction term, which stabilizes the learning dynamics and makes the model agnostic to the chosen noise distribution. The denoiser is expressed as
\begin{equation}
D_\theta(\mathbf{x}, \sigma)
= 
c_{\text{skip}}(\sigma)\, \mathbf{x}
+ 
c_{\text{out}}(\sigma)\,
F_\theta\!\left(
c_{\text{in}}(\sigma)\, \mathbf{x},\, c_{\text{noise}}(\sigma)
\right),
\label{eq:edm_preconditioning}
\end{equation}
where $F_\theta$ is the raw neural network and $(c_{\text{in}}, c_{\text{out}}, c_{\text{skip}})$ are preconditioning coefficients that depend on the noise level $\sigma$:
\begin{align}
c_{\text{in}}(\sigma) &= \frac{1}{\sqrt{\sigma^2 + \sigma_{\text{data}}^2}}, \\
c_{\text{out}}(\sigma) &= \frac{\sigma \, \sigma_{\text{data}}}{\sqrt{\sigma^2 + \sigma_{\text{data}}^2}}, \\
c_{\text{skip}}(\sigma) &= \frac{\sigma_{\text{data}}^2}{\sigma^2 + \sigma_{\text{data}}^2}.
\end{align}
These scaling functions control the dynamic range of the network and ensure that the denoiser operates consistently at all noise levels, leading to faster convergence and improved fidelity.

The noise schedule $\sigma(t)$ defines how the noise level changes over time and strongly influences both the training stability and sample quality. In our work, we follow the same noise schedule as used in the GenCast model~\cite{Price2024GenCast}. The authors adapt this by defining a noise level parameterized through a bounded power-law schedule that better suits large-scale geophysical fields. In this formulation, $\sigma(t)$ evolves nonlinearly with the diffusion step $t$:
\begin{equation}
\sigma(t) =
\left(
\sigma_{\min}^{1/\rho}
+ t \, \big(\sigma_{\max}^{1/\rho} - \sigma_{\min}^{1/\rho}\big)
\right)^{\rho},
\label{eq:gencast_schedule}
\end{equation}
where $\rho$ controls the curvature of the schedule. This schedule distributes diffusion steps more densely at low $\sigma$, focusing the denoising effort on the region where small errors have the largest perceptual and physical impact. The resulting sampling trajectory is smoother and more stable for large spatial fields such as weather or fluid data.

The EDM framework unifies prior diffusion approaches, such as variance-preserving and variance-exploding formulations, and introduces principal design components that make diffusion models practical for high-resolution physical data. The weighting function $\lambda(\sigma)$ prevents numerical imbalance between scales, while the preconditioning terms $(c_{\text{in}}, c_{\text{out}}, c_{\text{skip}})$ stabilize gradient magnitudes during training. The logarithmic normal noise schedule allows efficient coverage of the noise spectrum with fewer diffusion steps. Together, these choices lead to improved sample quality and a significant reduction in the number of solver evaluations required for high-fidelity generation.

In the context of urban fluid flow modeling, these properties are essential. Flow fields exhibit strong multiscale variability, and their probabilistic reconstruction requires accurate modeling of both low- and high-variance structures. Building upon the EDM foundation, our model combines diffusion-based probabilistic generation with graph neural networks to capture spatially structured dependencies on complex urban geometries.

\subsection{Denoiser architecture}
\label{sec:denoiser_architecture}

\begin{figure}[t]
  \centering
  \includegraphics[width=\linewidth]{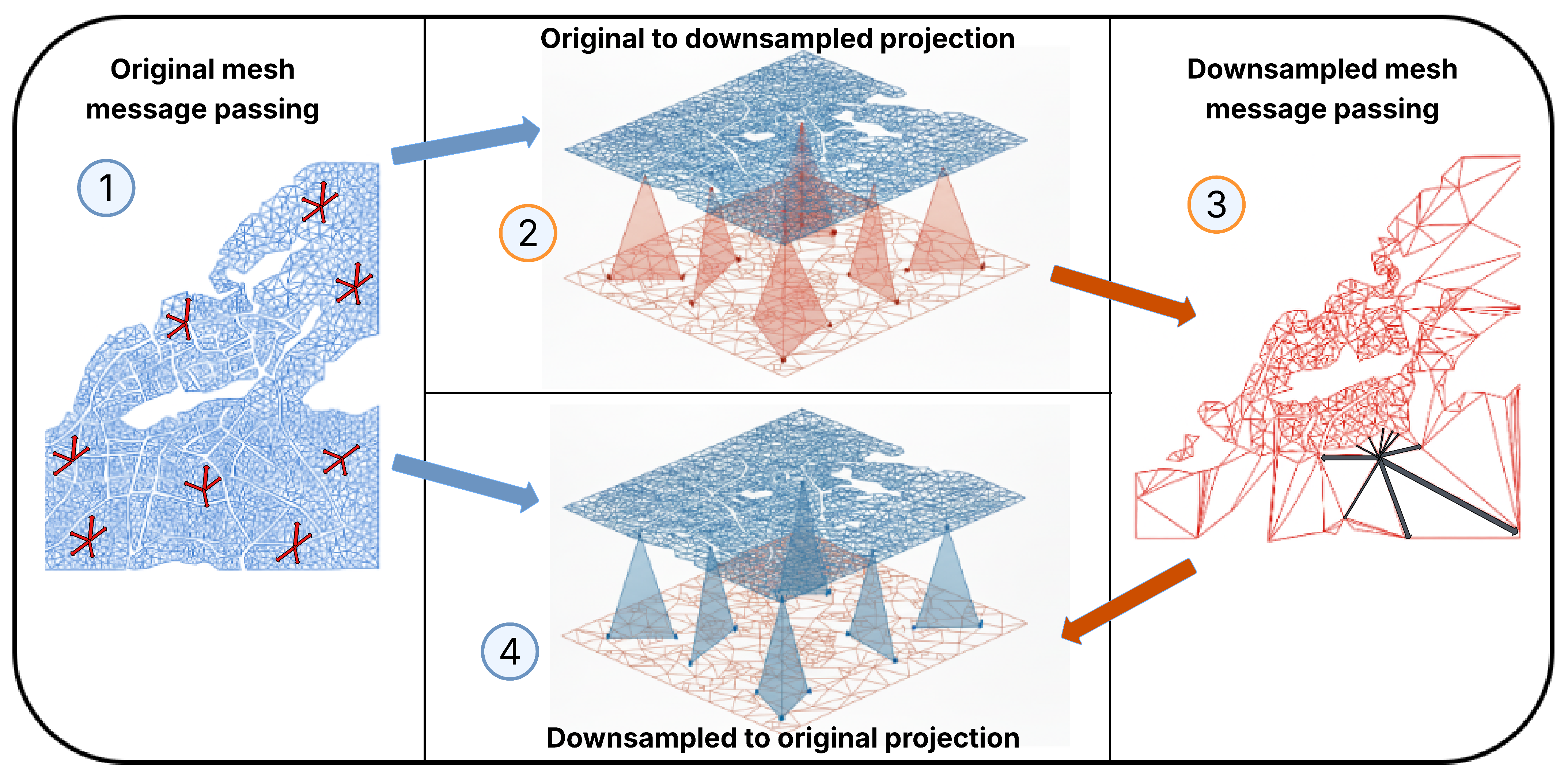}
  \caption{Hierarchical message passing on two meshes. 
  (1) Local updates on the original mesh, using only edges that connect nearby nodes in the high-resolution domain. 
  (2) Projection from the original mesh to a reduced mesh (downsampled). 
  (3) Long-range propagation on the reduced mesh, where fewer nodes allow information to travel efficiently across the entire domain. 
  (4) Projection back to the original mesh. 
  The reduced level accelerates non-local communication while predictions remain on the original unstructured domain.}
  \label{fig:architecture}
\end{figure}

\begin{figure}[t]
  \centering
  \includegraphics[scale=0.35]{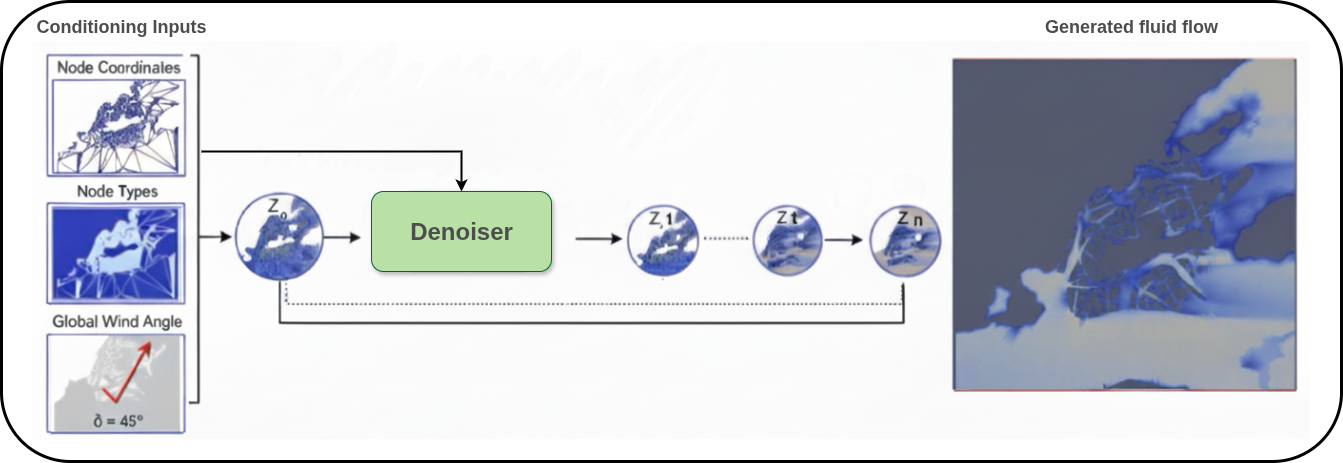}
  \caption{Overview of the diffusion denoising process. The denoiser receives noisy flow fields, node coordinates, node types, and the global wind angle as conditioning inputs, and predicts the denoised flow at each step.}
  \label{fig:diffusion_process}
\end{figure}

The present denoiser follows the multiscale design of MultiScale MeshGraphNets (MS--MGN)~\cite{fortunato2022msmgn} and the encode--process-decode structure used in MeshGraphNets~\cite{pfaff2021mgn} and GraphCast~\cite{lam2023graphcast}. It is composed of four encoder--processor--decoder GNNs, each operating on a different edge set that connects either the original mesh, the reduced mesh, or both. These four subnetworks enable information flow both within each resolution level and across levels, as illustrated in Figure~\ref{fig:architecture}. The networks correspond directly to the four steps shown in the figure: 
\begin{itemize}
    \item original $\rightarrow$ original (\texttt{o2o}): step (1) in Figure~\ref{fig:architecture}, performing local message passing on the original high-resolution mesh;
    \item original $\rightarrow$ reduced (\texttt{o2r}): step (2), projecting (downsampling) features from the original mesh to the reduced mesh; 
    \item reduced $\rightarrow$ reduced (\texttt{r2r}): step (3), transferring the message to the coarse mesh, enabling efficient long-range interactions; 
    \item reduced $\rightarrow$ original (\texttt{r2o}): step (4), projecting (upsampling) features from the reduced mesh back onto the original mesh.
\end{itemize}

Thus, \texttt{o2o}, \texttt{o2r}, \texttt{r2r}, and \texttt{r2o} refer to both the edge sets that define how nodes are connected across resolutions, and the corresponding GNNs that operate on each of these connectivity patterns.

Each subnetwork is implemented as a self-contained encoder--processor--decoder GNN. The encoder transforms raw node and edge features into latent embeddings, the processor performs iterative message passing over its associated edge set, and the decoder maps the processed latent states to the desired output representation. This modular design allows consistent graph operations across all four subnetworks while enabling multi-resolution communication similar to MS--MGN.

The network input consists of noisy velocity fields defined on the original mesh, together with node-level conditioning inputs. Each original-mesh node carries: (i) noise target channels, (ii) node-type one-hot encoding (wall, boundary, fluid), (iii) positional coordinates $(x,y)$, and (iv) directional features derived from the global wind angle, $d_\Phi =[\cos\Phi, \sin\Phi]$, broadcast to all nodes. The concatenation of these features yields $\mathbf{X}\in\mathbb{R}^{B\times N_o\times C_{in}}$.

Edge features include geometric and directional components. For each edge $i\!\to\!j$, we include the relative displacement $\mathbf{d}_{ij}=(dx,dy)$, the length of the edge $d=\|\mathbf{d}_{ij}\|$, and two projections aligned and perpendicular to the wind direction.
\[
p_{\mathrm{par}} = \frac{(\mathbf{x}_j-\mathbf{x}_i)\cdot\mathbf{d}}{d}, \qquad 
p_{\mathrm{perp}} = \frac{(\mathbf{x}_j-\mathbf{x}_i)\cdot\mathbf{d}_\perp}{d},
\]
where $\mathbf{d}=[\cos\Phi,\sin\Phi]$ and $\mathbf{d}_\perp=[-\sin\Phi,\cos\Phi]$.  
These edge features are defined separately for each edge set (\texttt{o2o}, \texttt{o2r}, \texttt{r2r}, \texttt{r2o}), matching the connectivity patterns used by the four subnetworks.

Figure~\ref{fig:diffusion_process} illustrates the entire diffusion process. The model takes as input node coordinates, node types, and the global wind direction $\Phi$, together with an initial noise field. Through multiple denoising steps, the model progressively reconstructs a coherent fluid flow field on the input mesh.

The global conditioning vector encodes both the diffusion noise level $\sigma$ and the wind direction $\Phi$. The noise level is embedded using a Fourier-features MLP, while the angle is embedded using a harmonic MLP. These encodings are fused into a single conditioning vector that modulates all message-passing blocks through normalization-based conditioning, allowing the network to adapt dynamically to both the diffusion step and external forcing conditions.

Each processor component may contain multiple message-passing layers: in our implementation, the \texttt{o2o} network uses two layers, the \texttt{r2r} network uses six layers, and the projection networks (\texttt{o2r}, \texttt{r2o}) use one layer each. Hierarchical propagation follows
\[
\texttt{o2o} \rightarrow \texttt{o2r} \rightarrow \texttt{r2r} \rightarrow \texttt{r2o},
\]
where \texttt{o2o} performs local interactions on the fine mesh (information exchange only between spatially adjacent CFD nodes), while \texttt{r2r} enables long-range interactions by propagating information across the entire domain on a much smaller graph. 
The final prediction $\widehat{\mathbf{Y}} \in \mathbb{R}^{B \times N_o \times C_{\text{out}}}$ is produced on the original mesh through the \texttt{r2o} decoder.

For additional details on the encode–process–decode structure and the conditioning mechanism, refer to Appendix~\ref{app:gnn_details}.

\subsection{Urban flows dataset}
\label{sec:dataset}

The dataset used in this work originates from high-fidelity CFD simulations of urban wind flow over a real 3D city geometry, in particular a neighborhood of Bristol, UK. The domain consists of an irregular terrain and detailed building structures enclosed within a cylindrical control volume. The 3D mesh contains approximately 40 million cells.

The flow data correspond to steady-state RANS simulations that provide the velocity components $(U_x, U_y, U_z)$ for 360 angles of attack of the incoming wind. Each case represents a converged stationary solution under a different wind direction.

Urban wind flows are governed by the incompressible Navier–Stokes equations:
\begin{equation}
\nabla \cdot \mathbf{u} = 0,
\end{equation}
\begin{equation}
\rho \left( \mathbf{u} \cdot \nabla \mathbf{u} \right) 
= - \nabla p + \mu \nabla^2 \mathbf{u} + \mathbf{f},
\end{equation}
where $\rho$ denotes air density, $\mu$ is dynamic viscosity, and $\mathbf{f}$ accounts for body forces such as buoyancy or Coriolis effects. 

Steady-state RANS formulations are commonly used, decomposing the velocity field into mean and fluctuating parts, $\mathbf{u} = \overline{\mathbf{u}} + \mathbf{u}'$, which yields:
\begin{equation}
\rho \left( \overline{\mathbf{u}} \cdot \nabla \overline{\mathbf{u}} \right)
= - \nabla \overline{p}
+ \mu \nabla^2 \overline{\mathbf{u}}
- \nabla \cdot \left( \rho \overline{\mathbf{u}'} \right).
\end{equation}
The additional term 
$\rho \overline{\mathbf{u}' \mathbf{u}'}$ 
is the Reynolds stress tensor, approximated using turbulence closure models such as $k$--$\varepsilon$ or $k$--$\omega$. Solving these equations implies  meshes of tens or even hundreds of millions of cells for hundreds of wind directions (as in urban comfort assessment or dispersion analysis) incurs extreme computational cost, motivating surrogate models that can rapidly generate physically plausible mean flows without running CFD repeatedly.

To adapt the dataset to our 2D diffusion model, we extract multiple horizontal slices of the 3D flow field at different altitudes, obtaining planform views of the city. Each slice produces a distinct 2D mesh due to the irregular topography and varying building heights. This setup enables evaluation of the model’s generalization across geometries not seen during training.

We select six altitude slices ($z\!\in\!\{15,20,28,35,40,45\}$\,m), cropping the domain to $[-1000,1000]\!\times\![-1000,1000]$ and decimating to maintain a comparable number of nodes per mesh ($\sim$300.000 and $\sim$1.6-1.7 million edges on the  input (\texttt{o2o}) mesh). Four slices are used for training and two for testing. For each slice, we construct both the original and reduced meshes and define the four edge sets (\texttt{o2o}, \texttt{o2r}, \texttt{r2r}, \texttt{r2o}) following the multiscale graph procedure described in Section~\ref{sec:denoiser_architecture}. Nodes are labeled as fluid, wall, or boundary to provide explicit conditioning on solid and inflow regions.

\begin{figure}[t]
  \centering
  \includegraphics[width=\linewidth]{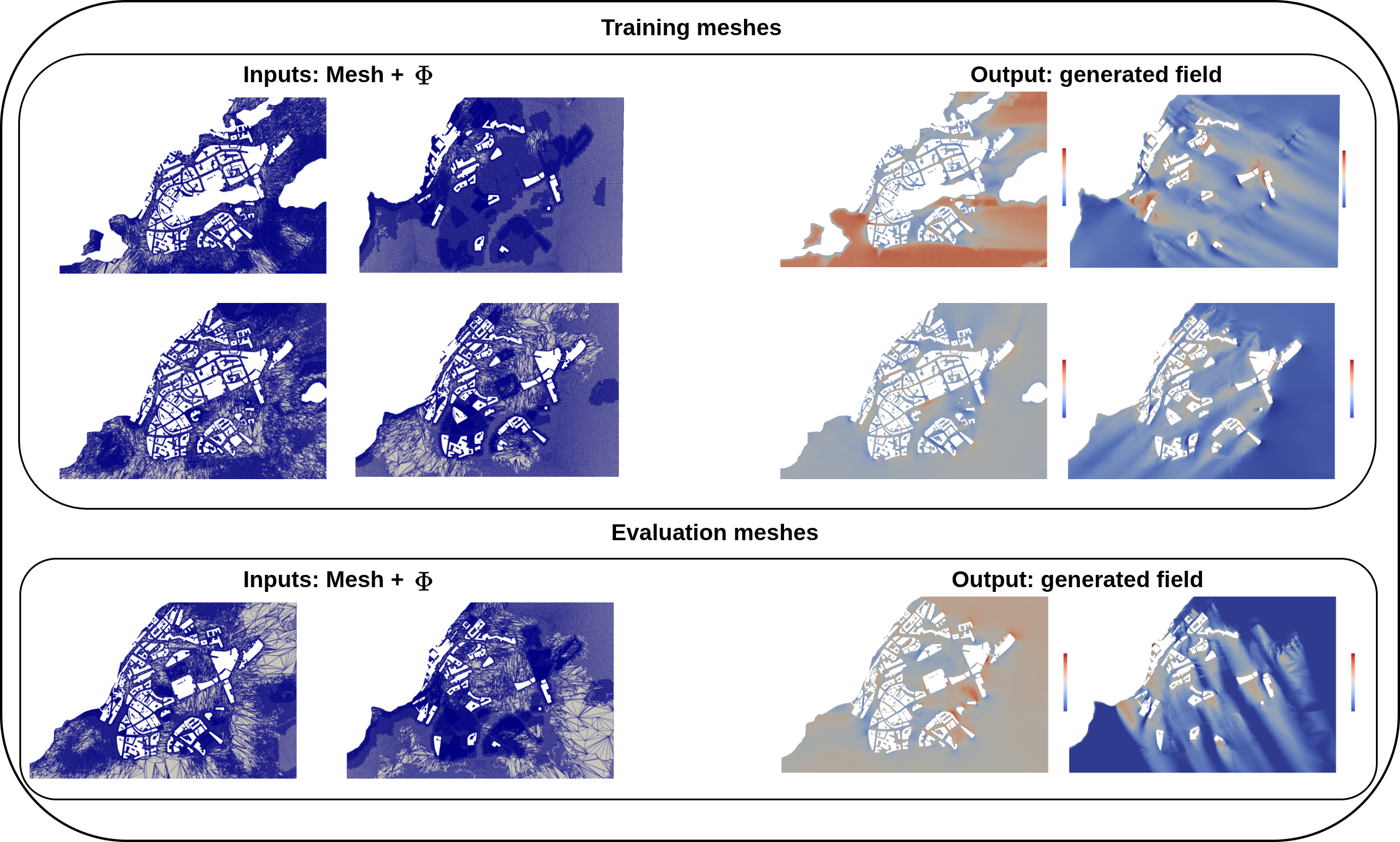}
  \caption{Training and evaluation meshes differ in both obstacle configuration and mesh resolution. The model receives only the mesh and inflow wind direction ($\Phi$) as input and generates the corresponding flow field. On the right, we observe how the resulting flow patterns change with the prescribed wind direction; each mesh includes 360 possible inflow angles, allowing the model to learn and generalize across a wide range of inflow conditions.}
  \label{fig:training_and_test_data}
\end{figure}

The separation between training and testing meshes is shown in Figure~\ref{fig:training_and_test_data}. Notice that the geometries differ significantly across meshes, including both the number and spatial arrangement of buildings, as well as the density and distribution of mesh nodes. In our formulation, the input to the model is solely the mesh information (node coordinates, connectivity, and node types), and the output is the generated velocity field conditioned on the corresponding geometry.

More details, mesh visualizations, and dataset statistics are provided in the Appendix~\ref{app:dataset_details}.

\subsection{Graph diffusion urban fluid flow fields generation}

We formulate urban wind flow generation as a conditional probabilistic modeling task in which, given a specific urban geometry (represented as a mesh) and a global wind direction, the goal is to sample physically consistent 2D velocity fields. Unlike supervised learning approaches that attempt to regress a single deterministic ground-truth field from input conditions, our objective is to learn the full conditional distribution
\[
p(\mathbf{u} \mid \Phi, \mathcal{G}),
\]
where $\mathbf{u}$ denotes the nodewise 2D velocity field, $\Phi$ is the incoming wind angle, and $\mathcal{G}$ encodes the geometry via node coordinates, connectivity, and boundary/wall labels. A direct supervised regression model would need to learn a highly non-local and implicitly constrained mapping because the velocity field must satisfy global incompressibility, momentum balance, and boundary conditions simultaneously, constraints that cannot be inferred from local geometry alone. Moreover, rather than being determined purely by geometric features, the flow must correspond to a solution of the steady Navier–Stokes equations for the given boundary conditions, which imposes additional physical structure on the mapping. As a result, a feedforward predictor tends to memorize training geometries rather than infer the underlying physical relationships, leading to poor generalization on new meshes.

In steady RANS simulations, the solution is obtained through an iterative fixed-point procedure that enforces global equilibrium between advection, diffusion, pressure forces, and wall constraints. This means that the converged field emerges only after repeated global updates, not from a single-pass mapping from geometry to velocity. Diffusion-based generation addresses this limitation by framing flow synthesis as an iterative refinement procedure. The denoising process progressively reconstructs the velocity field from noise in multiple steps, which function analogously to solver iterations. Each diffusion step updates the entire field while taking into account global context and local constraints, enabling the model to approximate the same equilibrium-seeking behavior that characterizes steady-state CFD solvers. This iterative structure allows the model to enforce global and local flow consistency over the mesh, rather than predicting all velocities in a single forward pass.

To address this, we employ a graph-conditioned diffusion process, where sampling begins from an isotropic Gaussian field defined over the mesh nodes and progressively evolves toward a physically coherent flow configuration. The reverse diffusion process is parameterized by the MS--MGN (described in Section ~\ref{sec:denoiser_architecture}), which is well-suited to unstructured meshes and can propagate information across complex topologies. At each denoising step, the MS--MGN takes as input the current noisy velocity, the features of the nodes (coordinates, type of nodes) and the direction of the wind $\Phi$, and outputs a denoised prediction consistent with the EDM formulation described in Section~\ref{sec:edm_background}.

This probabilistic graph-diffusion formulation addresses several limitations observed in other learning approaches:

\begin{itemize}
    \item \textbf{Grid-based deep learning models} (e.g., CNNs or U-Net–based architectures) require structured, uniformly spaced grids. Applying them to urban meshes necessitates interpolation or voxelization, which can distort building boundaries, erode fine-scale shear regions, and introduce artifacts in recirculation zones.
    
    \item \textbf{Deterministic supervised predictors} trained to directly regress $\mathbf{u}$ from $(\Phi,\mathcal{G})$ approximate only the conditional mean of the distribution, which leads to excessive smoothing in regions of high variability such as separation layers, corner vortices, or wake transitions.
    
    \item \textbf{Autoregressive generative models} require a causal ordering or sequential factorization (e.g., temporal or spatial progression). In our setting, there is no natural generation order over nodes, nor a meaningful initial condition from which to autoregress; arbitrarily imposing an ordering risks introducing directional bias and breaking symmetry.
\end{itemize}

In contrast, diffusion-based generation with GNN denoisers produces coherent flow fields by iteratively refining noisy inputs while respecting both global context (via conditioning on $\Phi$ and global message passing) and local constraints (via mesh connectivity and boundary awareness). The iterative refinement allows the network to incrementally restore structures at different scales, which aligns naturally with the hierarchical organization of turbulent and urban flow features.

We also emphasize that using a graph representation enables the model to generalize to new geometries without retraining, provided that node features and connectivity are constructed similarly. The diffusion process thus becomes a flexible, mesh-aware, probabilistic generator that can synthesize urban wind patterns efficiently, bypassing the cost of repeated CFD evaluations.

\section{Results}
\label{sec:results}

We evaluated the proposed diffusion--GNN model in two unseen geometry slices (z=35 m and z=40 m), as defined in Figure \ref{fig:training_and_test_data}, in all test wind directions. Unless otherwise stated, all quantitative metrics are computed per angle and then aggregated across angles as mean~$\pm$ standard deviation (std). Our analysis first demonstrates the model's ability to capture spatial structure, distributional properties, and flow statistics. We then study the impact of the number of sampling steps on accuracy and cost, formalize the evaluation metrics used, and conclude with an ablation contrasting our multiscale architecture against a single-mesh baseline.

\begin{figure*}[t]
  \centering
  \includegraphics[width=0.97\linewidth]{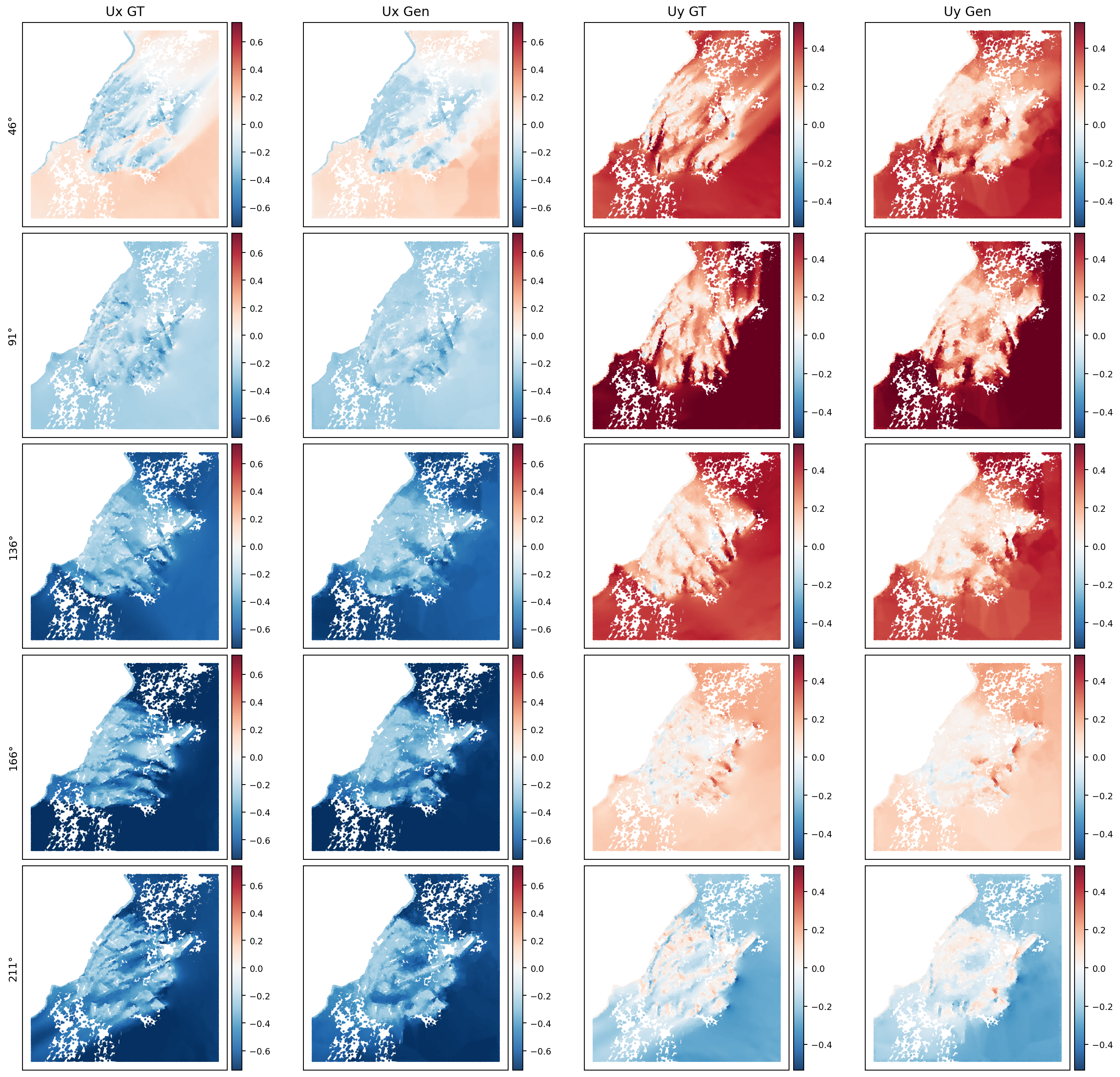}
  \caption{Component-wise velocity comparison across angles at $z{=}35$\,m. Ground-truth (left) vs.\ generated (right) fields for $U_x$ (left and center-left) and $U_y$ (center-right and right). }
  \label{fig:uxuy_compare}
\end{figure*}

To assess how well the model captures realistic spatial wind patterns, we show in Figure~\ref{fig:uxuy_compare} full-slice comparisons of ground-truth vs.\ generated velocity components ($U_x$, $U_y$) across representative angles at $z{=}35$\,m. The model accurately captures the dominant inflow direction, shear gradients near building edges, and recirculation zones downstream of urban obstacles. Fine-scale variation is also preserved, particularly at mid-to-high angles (e.g., $135^\circ$, $166^\circ$), where complex wake structures emerge.

\begin{figure*}[t]
  \centering
  \begin{subfigure}[t]{0.65\linewidth}
    \includegraphics[width=\linewidth]{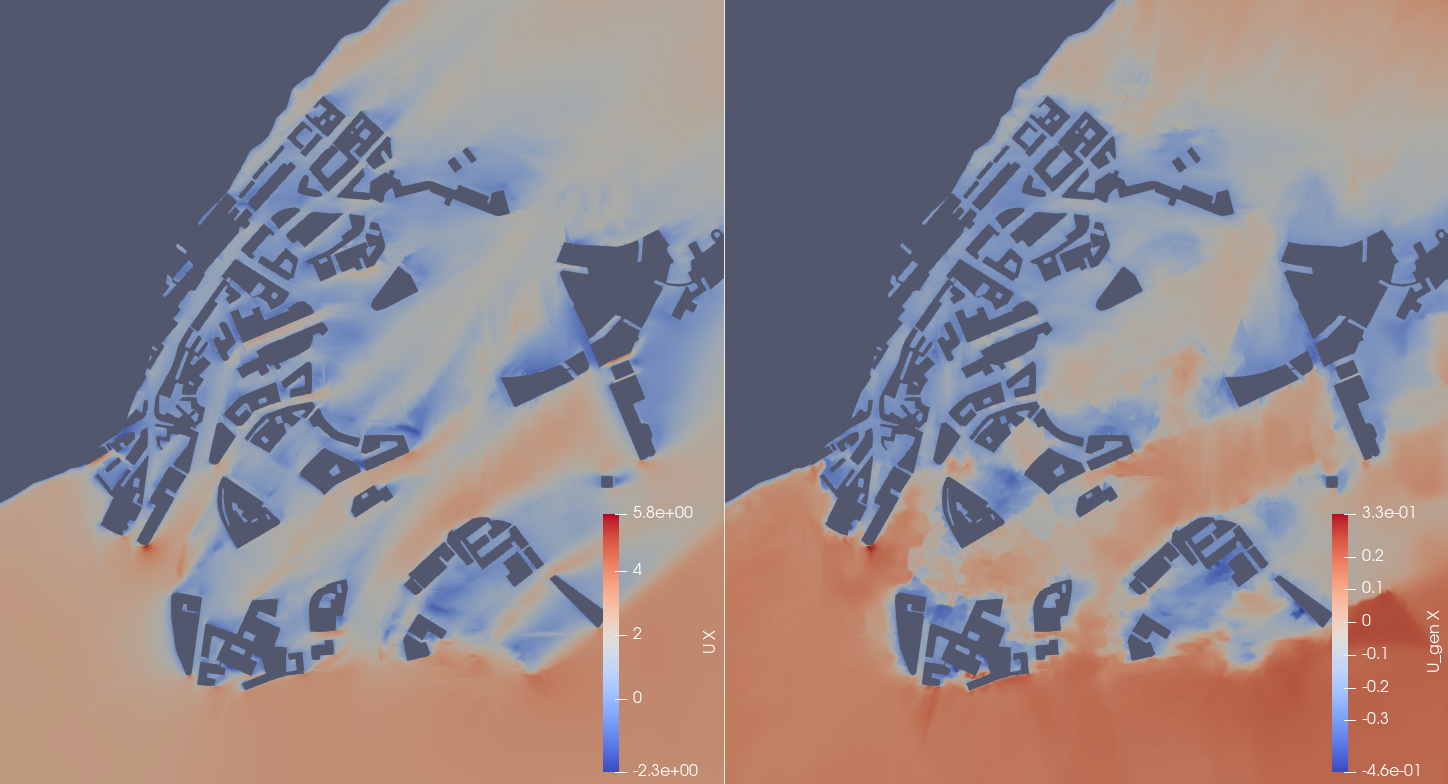}
    \caption{}
  \end{subfigure}\hfill
  \begin{subfigure}[t]{0.65\linewidth}
    \includegraphics[width=\linewidth]{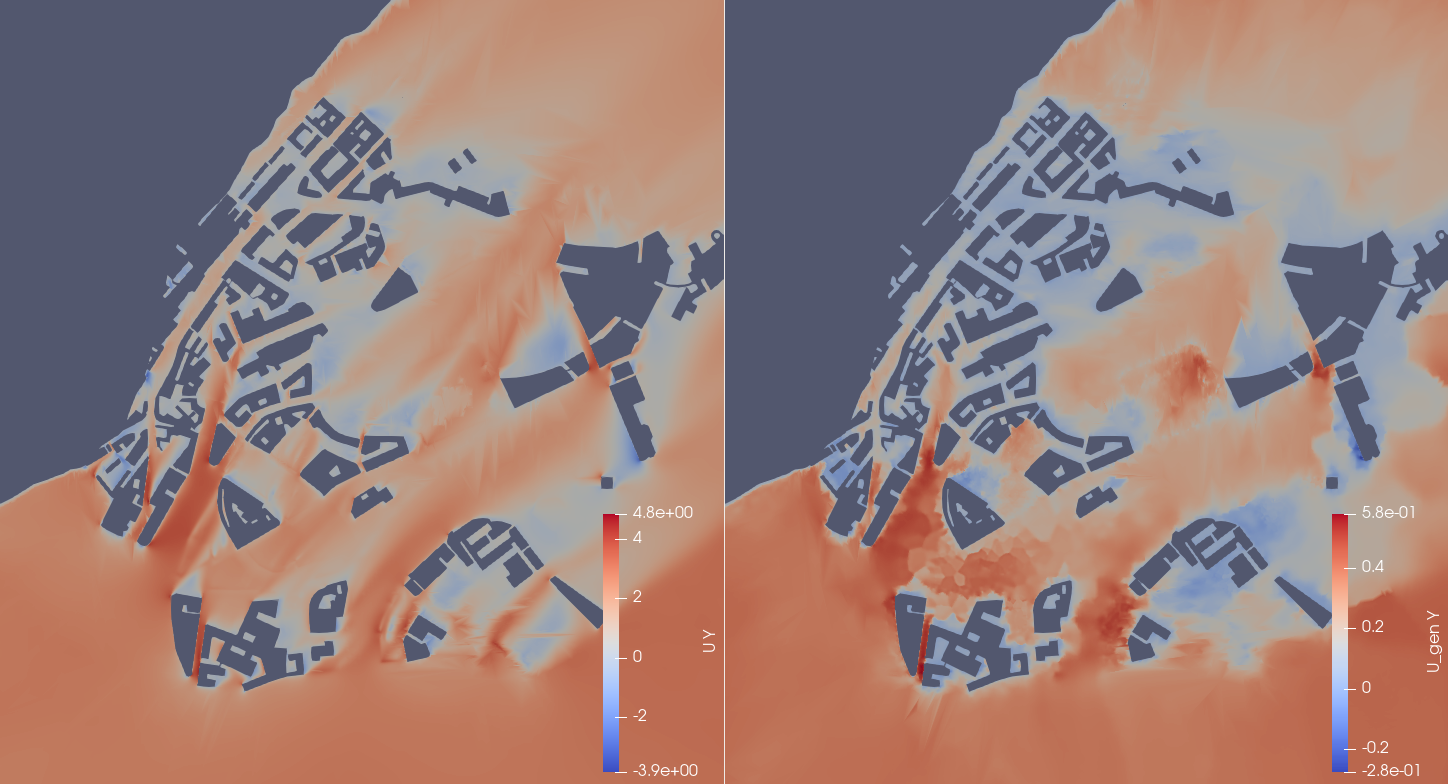}
    \caption{}
  \end{subfigure}
  \caption{Zoomed velocity components evaluated over a test slice, angle $46^\circ$. (a) $U_x$ component (zoomed view). Ground-truth (left) vs.\ generated (right). (b) $U_y$ component (zoomed view). Ground-truth (left) vs.\ generated (right)}
  \label{fig:zoom_fields}
\end{figure*}

To further assess flow structure at higher resolution, Figure~\ref{fig:zoom_fields} shows zoomed-in $U_x$ and $U_y$ visualizations for a dense urban core. The generated fields match both the spatial organization and sign pattern of the ground truth, with coherent velocity decay in canyon interiors and strong gradients at rooftop edges.

\begin{figure*}[t]
  \centering
  \includegraphics[width=0.42\linewidth]{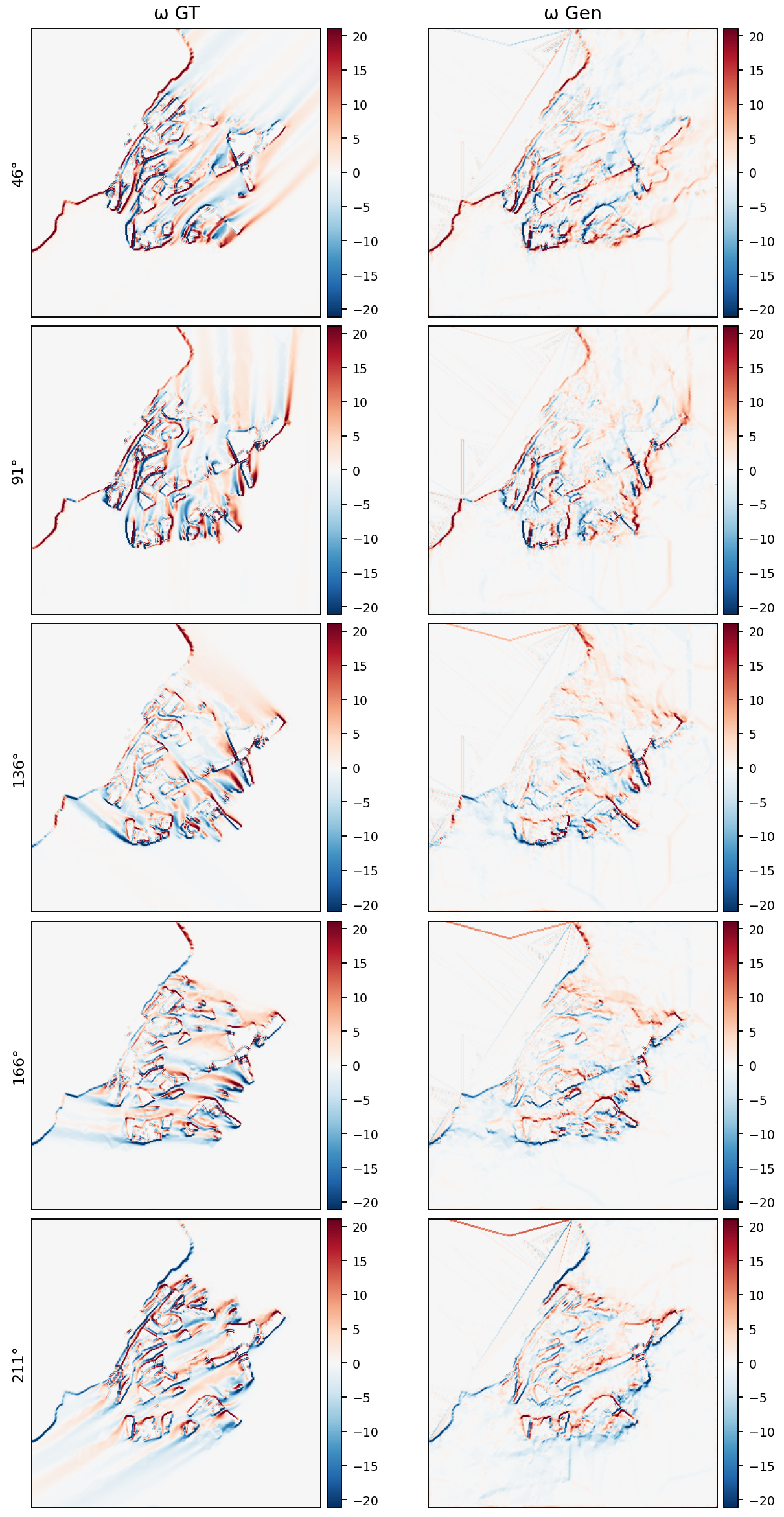}
  \caption{Vorticity comparison across angles at $z{=}35$\,m. Model output (right) preserves the spatial organization and peak locations of vorticity patterns seen in ground truth (left), though high-shear zones are mildly smoothed.}
  \label{fig:vorticity_compare}
\end{figure*}

Finally, we examine vorticity $\omega = \nabla \times \mathbf{u}$ in Figure~\ref{fig:vorticity_compare} to evaluate how well rotational features are reproduced. The generator preserves high-vorticity zones near sharp corners and flow separations, though mild smoothing is evident in regions of intense shear. These observations reinforce that the model not only approximates average flow but also captures dynamically relevant derivatives.

To evaluate whether the generated fields reproduce the statistical structure of the ground-truth flow, we compare their probability density functions (PDFs) for both velocity magnitude ($|\mathbf{u}|$) and components ($U_x$, $U_y$). For each field, we estimate the PDFs using kernel density estimation (KDE), a non-parametric method that approximates a continuous probability distribution from discrete samples \cite{silverman2018density, molinaro2024generative}. In KDE, each data value contributes a smoothly weighted local kernel (here a Gaussian), and the final density is obtained by summing these contributions across all nodes in the mesh. This approach avoids discretization artifacts that can arise from histogram binning and allows for consistent comparison between geometries with different mesh sizes.

\begin{figure*}[t]
  \centering
  \begin{subfigure}[t]{0.35\linewidth}
    \includegraphics[width=\linewidth]{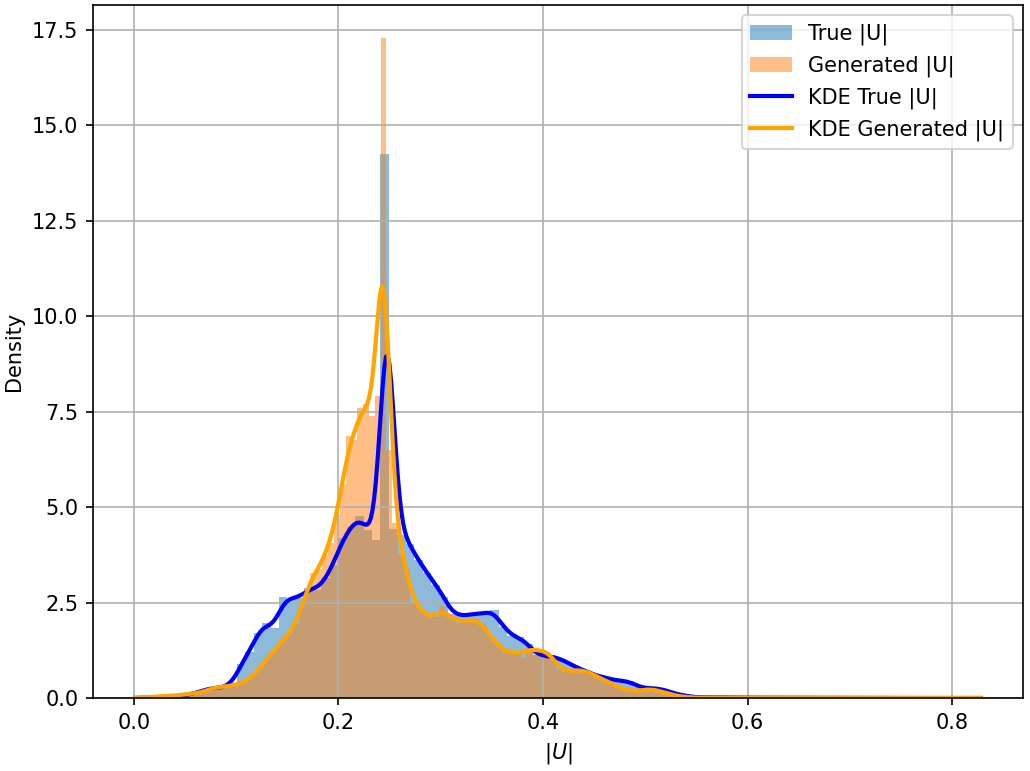}
    \caption{$|\mathbf{u}|$ ($z{=}40$, angle 31).}
  \end{subfigure}
  \begin{subfigure}[t]{0.35\linewidth}
    \includegraphics[width=\linewidth]{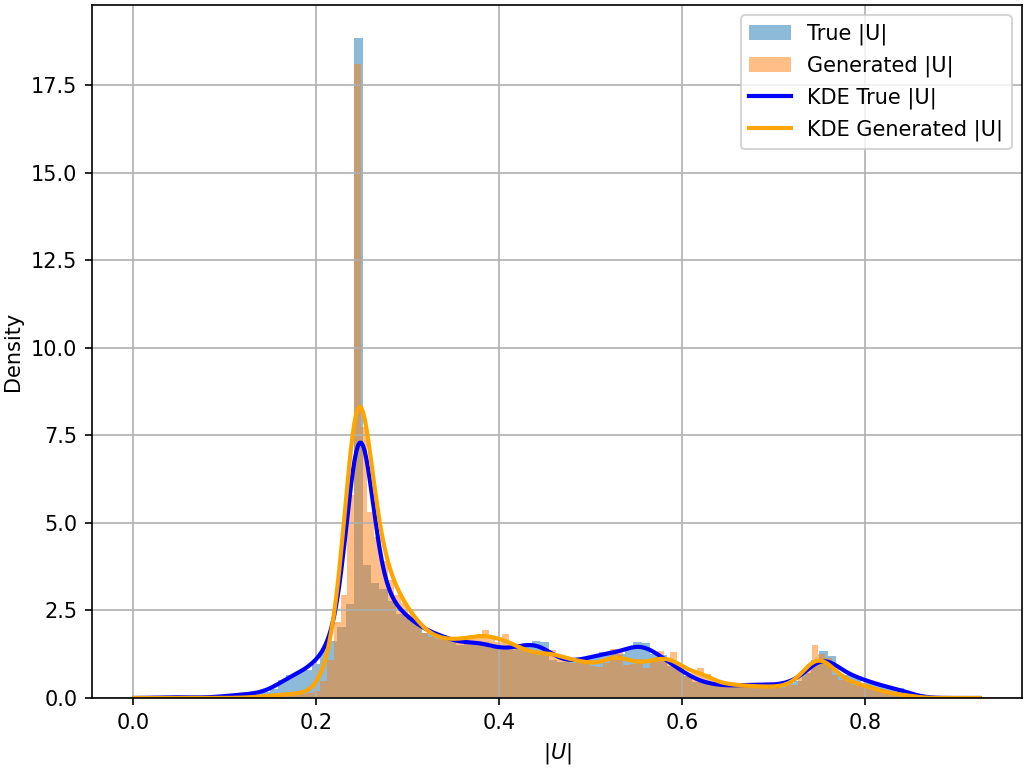}
    \caption{$|\mathbf{u}|$ ($z{=}35$, angle 211).}
  \end{subfigure}

  \begin{subfigure}[t]{0.35\linewidth}
    \includegraphics[width=\linewidth]{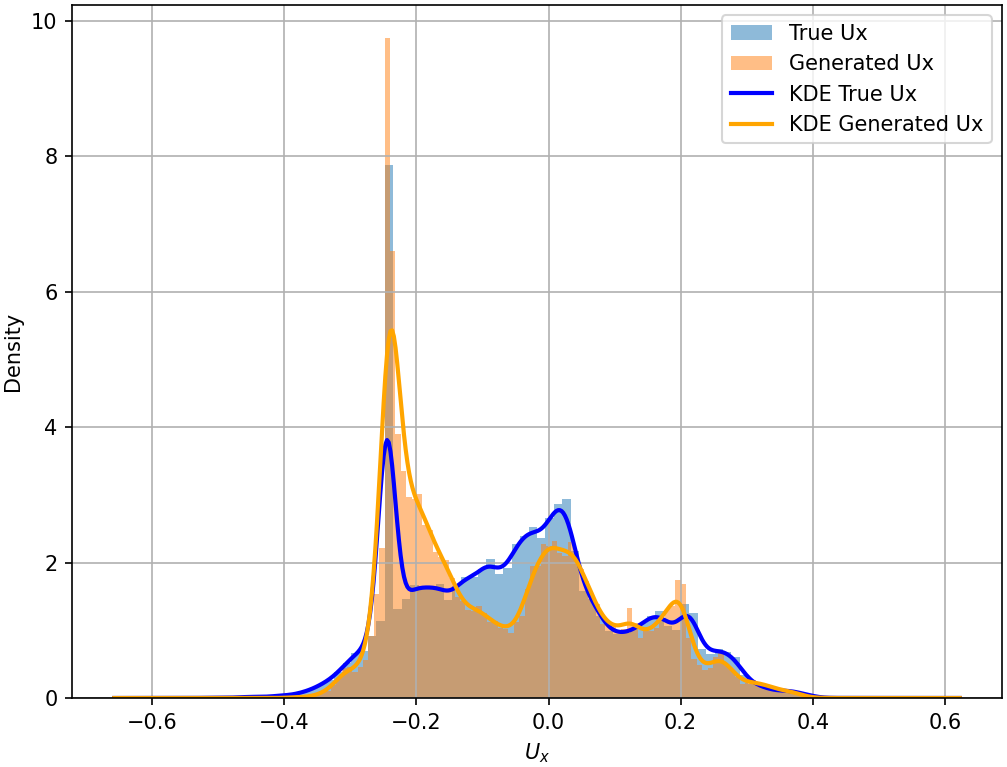}
    \caption{$U_x$ ($z{=}40$, angle 31).}
  \end{subfigure}
  \begin{subfigure}[t]{0.35\linewidth}
    \includegraphics[width=\linewidth]{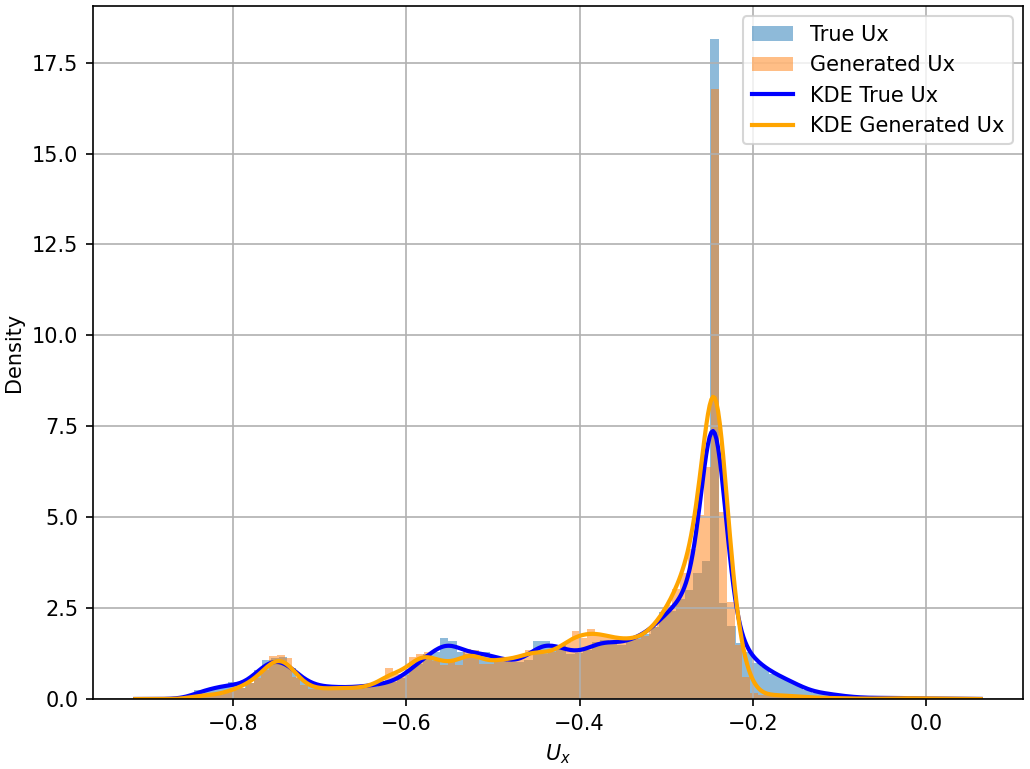}
    \caption{$U_x$ ($z{=}35$, angle 211).}
  \end{subfigure}

  \begin{subfigure}[t]{0.35\linewidth}
    \includegraphics[width=\linewidth]{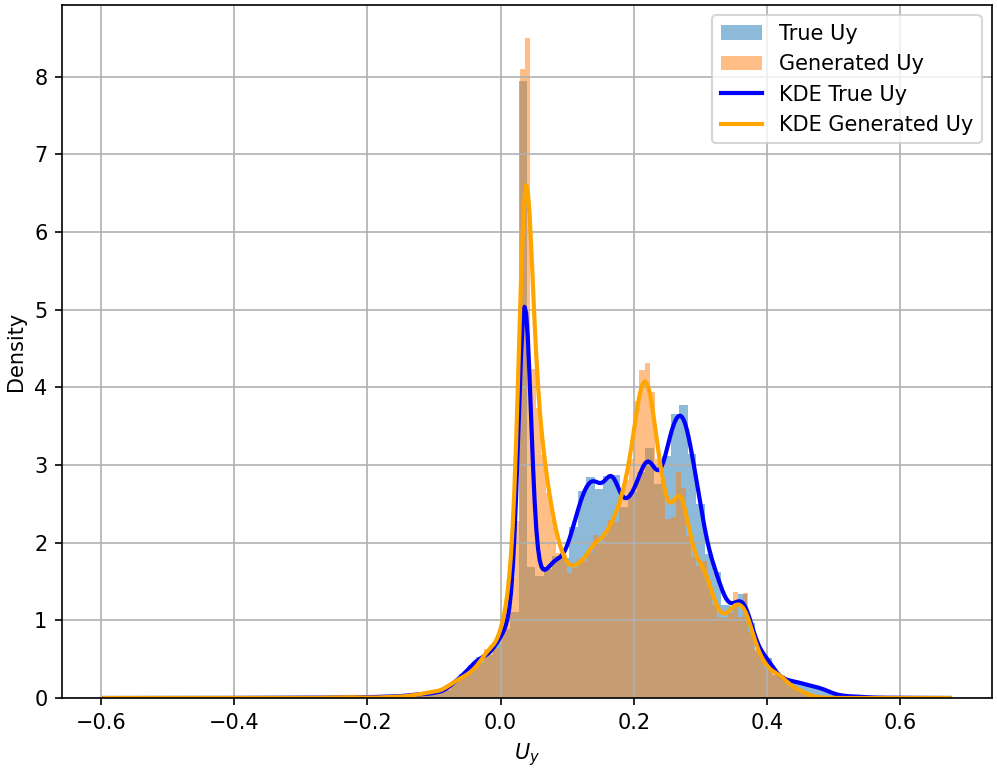}
    \caption{$U_y$ ($z{=}40$, angle 31).}
  \end{subfigure}
  \begin{subfigure}[t]{0.35\linewidth}
    \includegraphics[width=\linewidth]{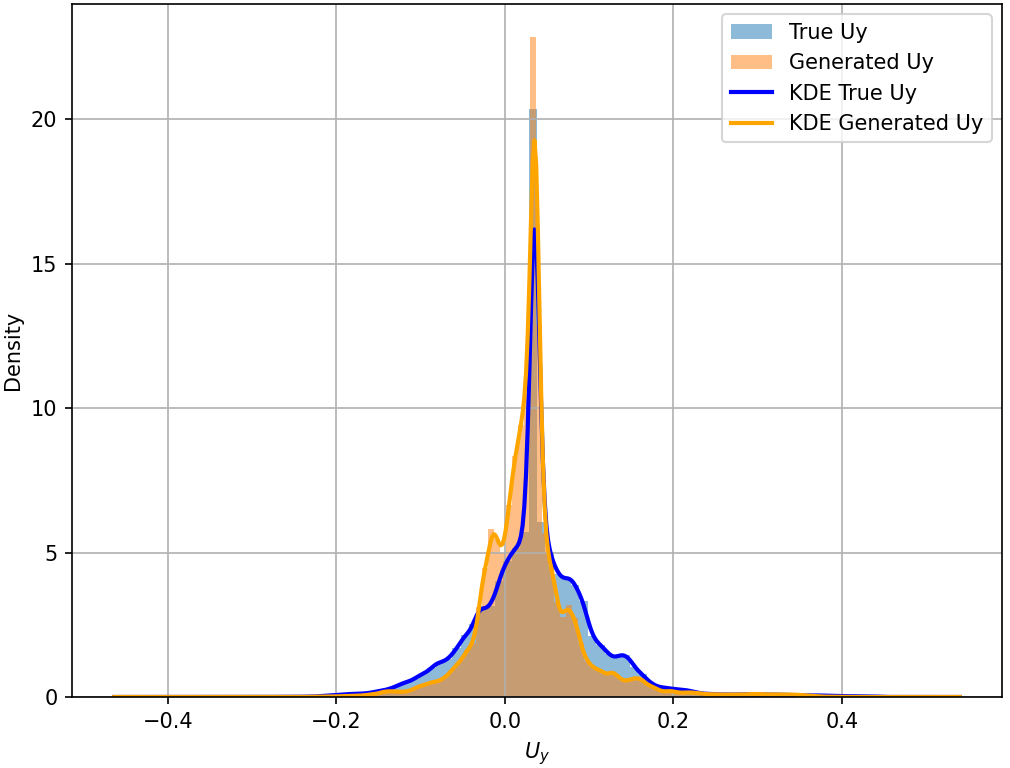}
    \caption{$U_y$ ($z{=}35$, angle 211).}
  \end{subfigure}
  \caption{PDF comparison for two representative angles at different heights. Predicted distributions track ground-truth statistics across components and speed; only slight tail underestimation is observed.}
  \label{fig:pdf_comparisons}
\end{figure*}

Figure~\ref{fig:pdf_comparisons} shows KDE-based PDFs for two representative inflow angles and altitudes. Across all variables, generated distributions closely match ground-truth bulk statistics and modal structure, which is quantitatively reflected by small 1-Wasserstein (Earth mover’s) distances \cite{panaretos2019statistical} between the KDE-based PDFs of generated and ground-truth fields. Averaged across all test cases, these distances are $0.012 \pm 0.008$ for $U_x$, $0.011 \pm 0.007$ for $U_y$, and $0.008 \pm 0.008$ for $|\mathbf{u}|$. These low distances indicate that discrepancies are limited to mild peak sharpening and a slight underrepresentation of extreme tails, consistent with moderate smoothing of high-gradient regions. In general, the statistical agreement suggests that the diffusion–GNN model preserves the probabilistic structure of urban wind fields while maintaining physical coherence.

\begin{figure*}[t]
  \centering
  \begin{subfigure}[t]{0.42\linewidth}
    \includegraphics[width=\linewidth]{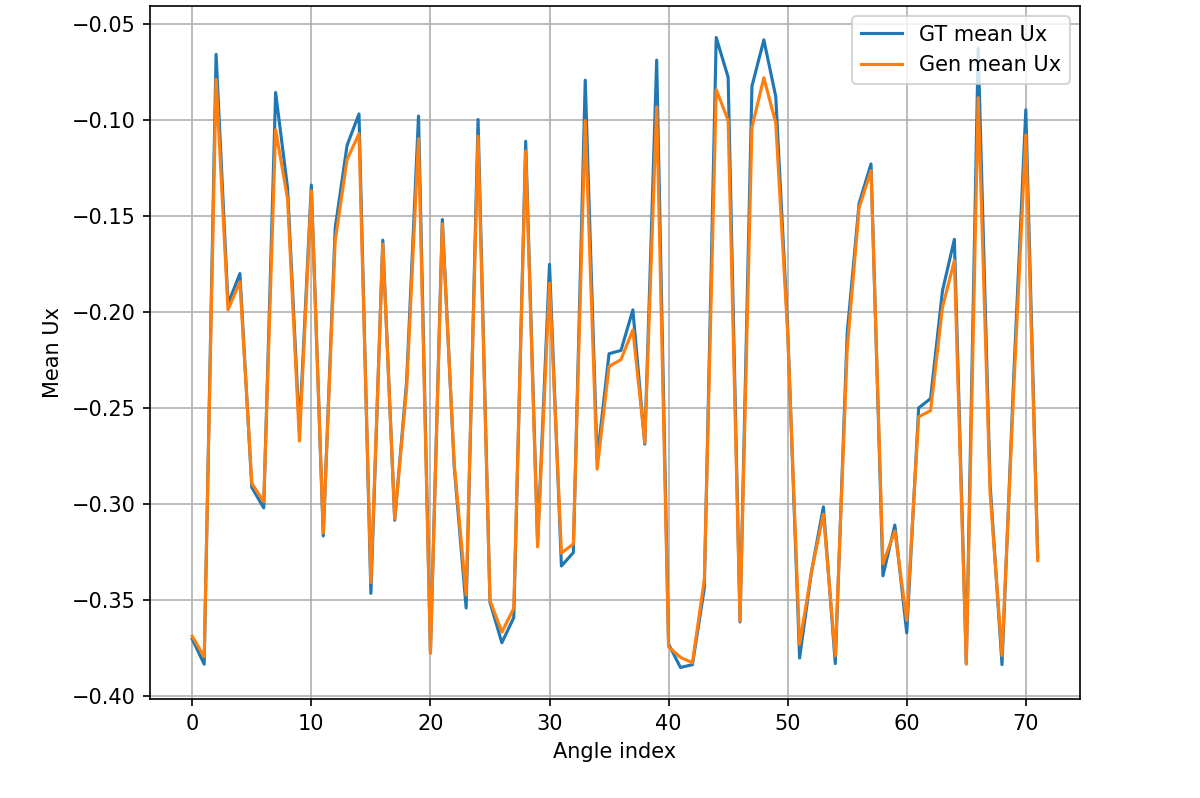}
    \caption{Mean $U_x$.}
  \end{subfigure}
  \begin{subfigure}[t]{0.42\linewidth}
    \includegraphics[width=\linewidth]{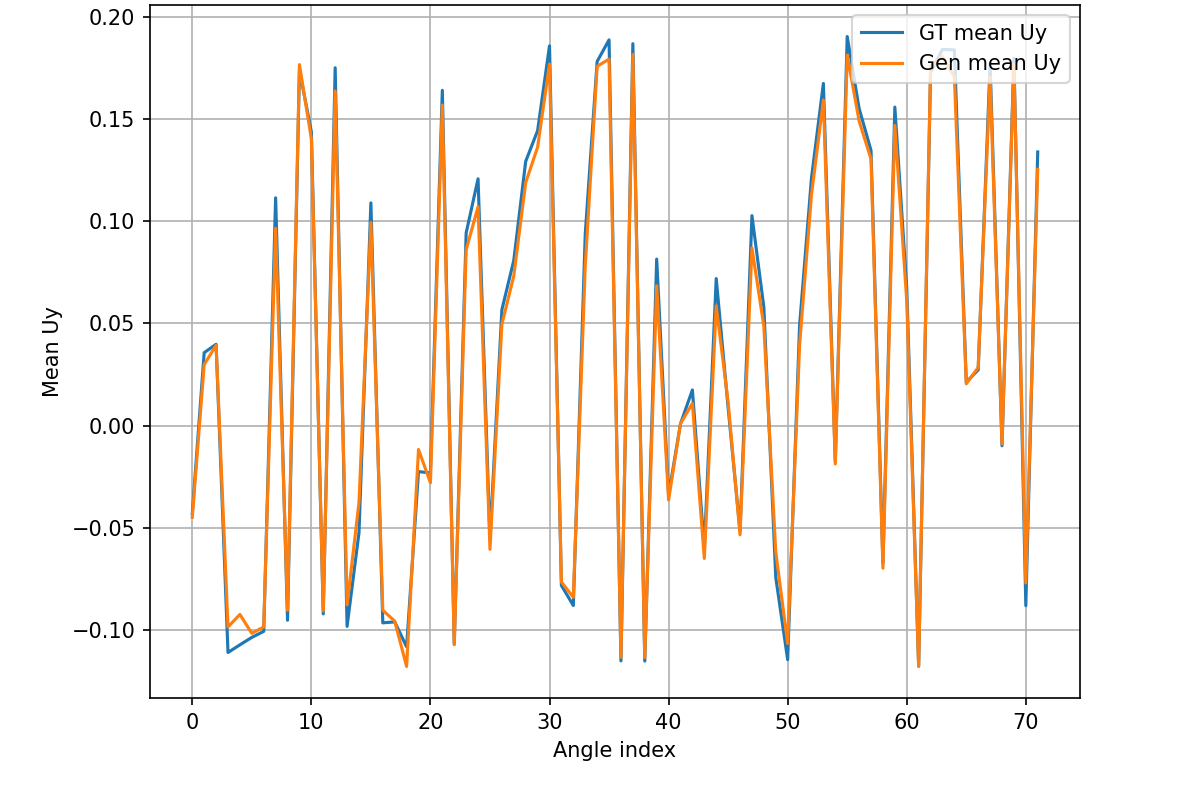}
    \caption{Mean $U_y$.}
  \end{subfigure}
  \caption{Angle-wise flow means at $z{=}35$\,m. The generator closely tracks GT variations across wind direction.}
  \label{fig:means_across_angles}
\end{figure*}

We next evaluate the model’s ability to reproduce angle-wise flow tendencies. 
Figure~\ref{fig:means_across_angles} shows the variation of angle-averaged speed and component means for the $z{=}35$ slice. 
The predicted trends closely follow GT behavior across angles, with deviations appearing mainly for angles aligned with long street canyons, where strong recirculation leads to sharper transitions.

To evaluate how well the model captures structures across spatial scales, we compute a Proper Orthogonal Decomposition (POD) of the velocity fields in the test set. POD provides an energy-ranked orthogonal basis in which the leading modes represent the dominant large-scale flow organization, while higher-order modes encode progressively finer-scale fluctuations \cite{zou2025generativeartificialintelligencehybrid}.

\begin{figure}[H]
  \centering
  \includegraphics[width=0.55\linewidth]{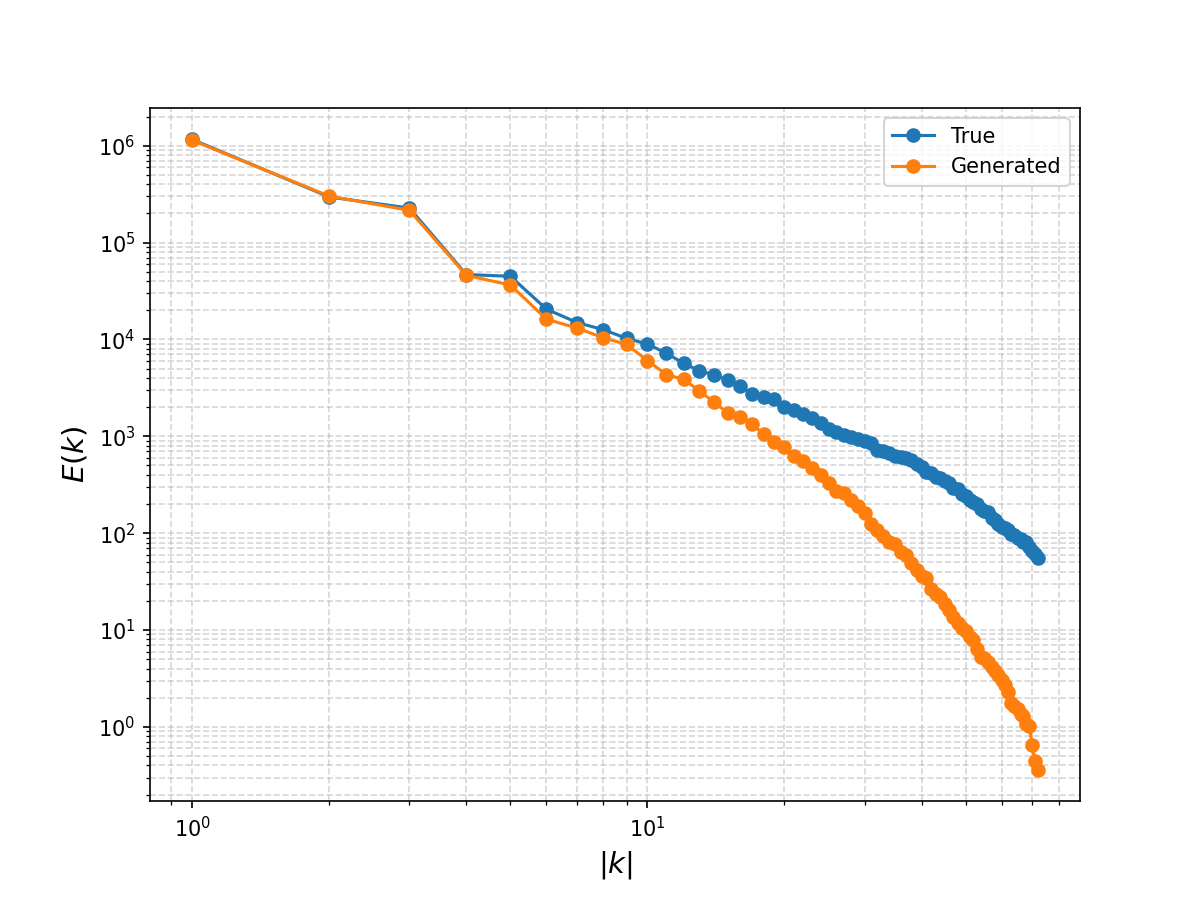}
  \caption{POD energy spectrum on the test set. The x-axis denotes the POD mode index (ordered by energy content). The y-axis shows the fraction of total kinetic energy captured by each mode.}
  \label{fig:pod_energy}
\end{figure}

As shown in Figure~\ref{fig:pod_energy}, the model accurately reproduces the leading POD energy content, indicating that the dominant coherent flow structures are well preserved. These leading modes correspond to the largest, most energetically significant patterns—such as primary recirculation regions, canyon-scale channeling, and wake organization—which govern the bulk dynamics of urban wind flow. The close agreement in these high-energy modes demonstrates that the generator retains the essential physical structure of the flow rather than merely matching local features.

Higher-order modes exhibit gradually reduced energy, reflecting mild smoothing of small-scale fluctuations. These lower-energy modes typically encode fine-scale spatial variations that can include turbulence-like fluctuations, numerical noise, redundant geometric details, or dynamically less relevant patterns. Their partial attenuation is therefore expected in a generative surrogate and consistent with the uncertainty and variability inherent to small-scale features in RANS solutions. Overall, the POD analysis confirms that the model captures the physically meaningful large-scale organization of the flow while only modestly smoothing structures that are intrinsically more variable or less critical to the governing dynamics.

The diffusion-GNN model successfully reproduces coherent spatial flow features, closely matching the statistical 
distributions of velocity magnitude and components across wind angles. KDE-based PDF comparisons demonstrate that the 
generated fields capture both the bulk statistics and the variability of the ground truth, reflecting accurate recovery of 
the underlying flow distribution. The POD energy spectra further show that the model preserves the dominant multiscale 
organization of the flow: the leading modes retain the correct kinetic-energy content, and the overall spectral decay is 
consistent with ground-truth behavior. Together, these results indicate that the model learns not only pointwise 
velocities but also their broader statistical and energetic structure, providing physically meaningful reconstructions 
across unseen geometries and inflow conditions.

\subsection{Sampler Steps vs Accuracy and Computational Cost}

\begin{figure*}[t]
  \centering
  \begin{subfigure}[t]{0.48\linewidth}
    \includegraphics[width=\linewidth]{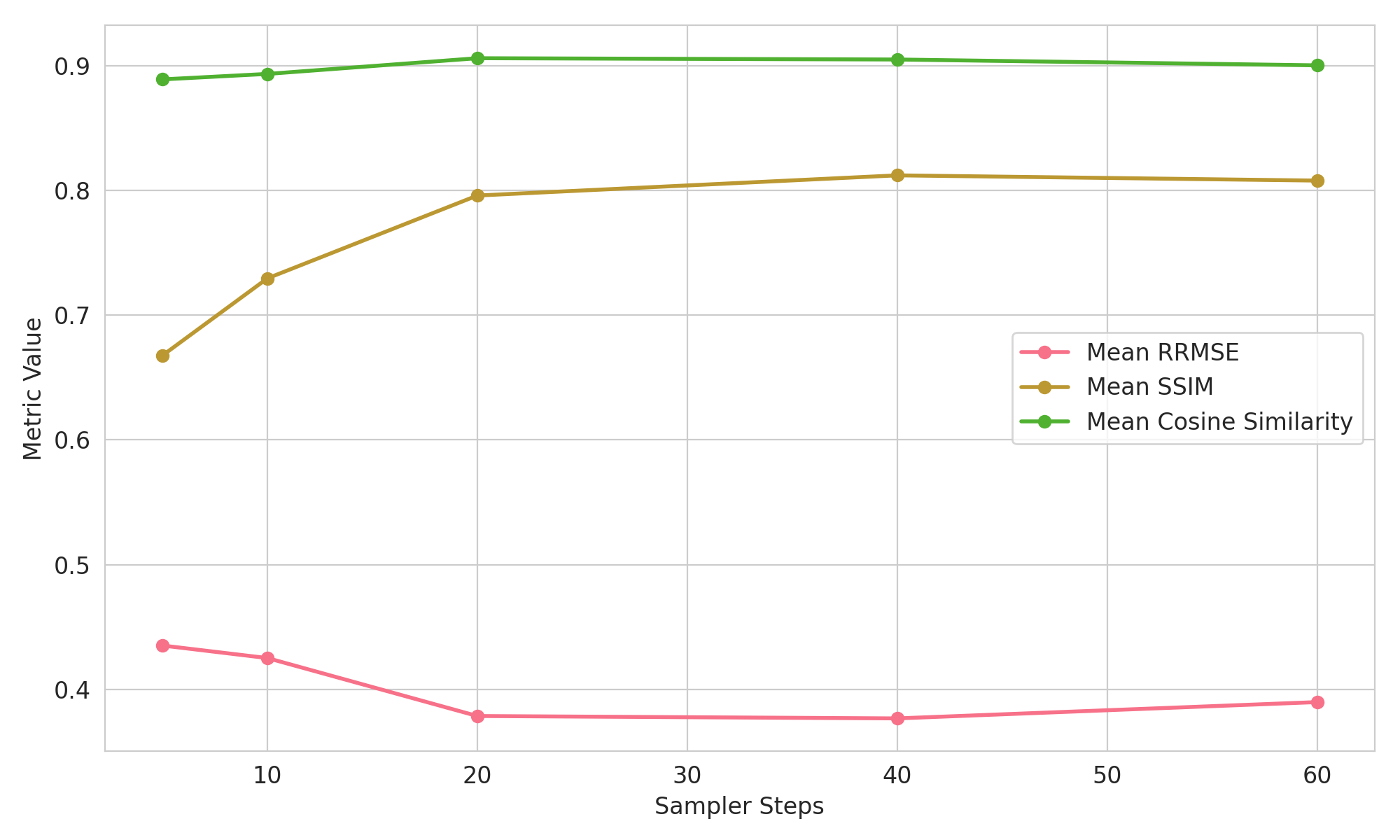}
    \caption{Metric evolution vs.\ steps (e.g., RMSE, SSIM, cosine).}
  \end{subfigure}\hfill
  \begin{subfigure}[t]{0.48\linewidth}
    \includegraphics[width=\linewidth]{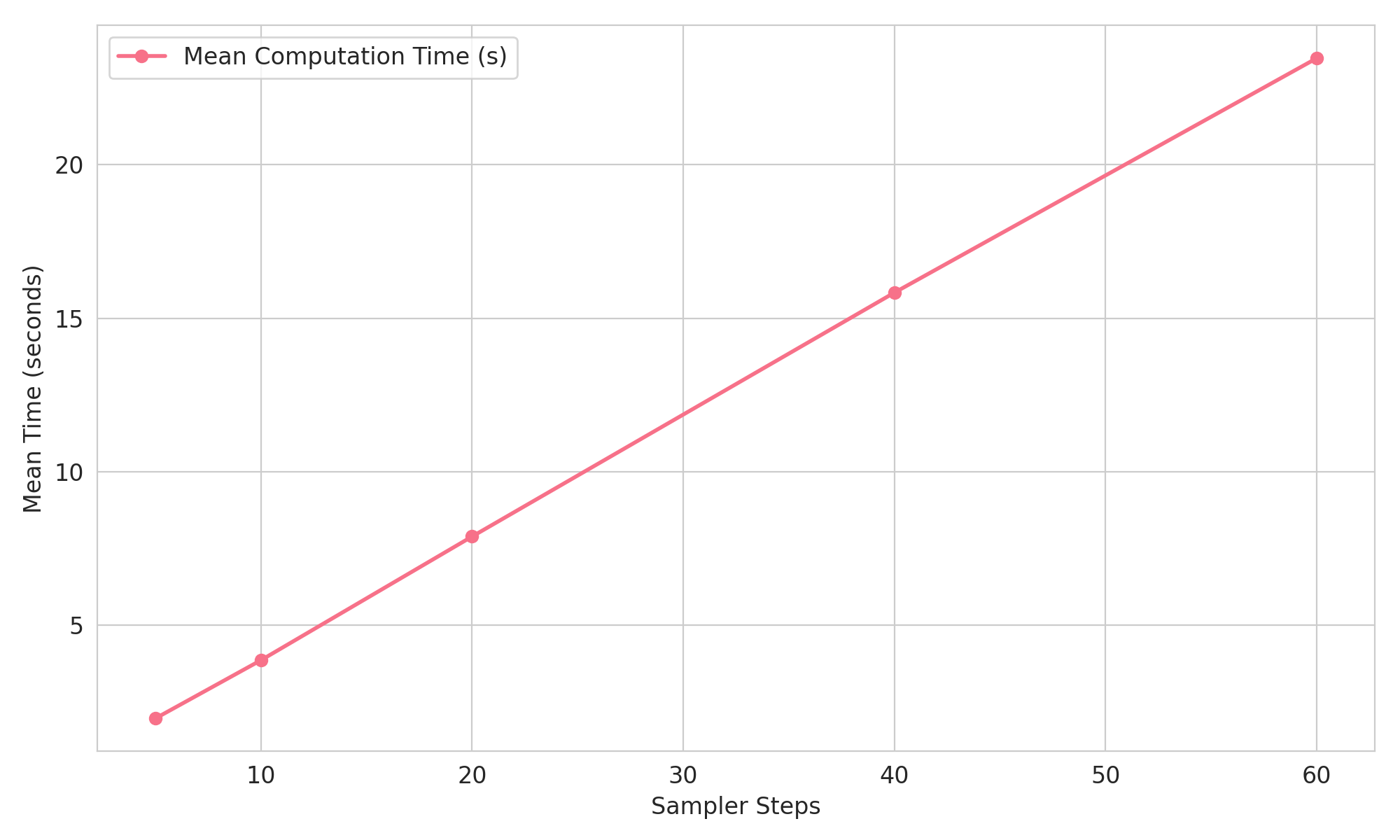}
    \caption{Mean inference time vs.\ steps.}
  \end{subfigure}
  \caption{Accuracy-cost trade-off with sampling steps. Quality saturates after $\sim$20--40 steps, while time scales linearly.}
  \label{fig:sampler_subplot}
\end{figure*}
We analyze how the number of sampling steps affects fidelity and runtime. 
Figure~\ref{fig:sampler_subplot} shows (left) the evolution of key metrics (e.g., RMSE, SSIM, cosine similarity) and (right) the corresponding mean wall-clock time. The definitions of all evaluation metrics used in the subsequent analysis are provided in Appendix~\ref{app:metrics}.

Accuracy improves rapidly between 5 and 20 steps, with modest gains up to $\sim$40 steps and little benefit beyond. 
Runtime increases nearly linearly. 
Therefore, 20--40 steps represent an effective trade-off between fidelity and efficiency. Selecting 20 denoising steps allows the generation of a case on ~7,5 seconds on a single A100 GPU.

Beyond the per-sample trends, we report the overall computational cost of training and inference.  The full model was trained for 8 hours using data parallelism across four NVIDIA A100\,40GB GPUs. Once trained, inference is extremely fast: generating all 360 wind directions for a given slice 
requires only $\sim$2700\,seconds on a single A100\,40GB (mean wall-clock time), corresponding to well under ten seconds per flow field. In comparison, producing the same set of RANS solutions requires multiple hours to days of computation on HPC infrastructure. This gap illustrates the practical value of the diffusion-GNN surrogate: steady-state urban wind fields can be synthesized almost in real time, enabling rapid scenario exploration and interactive design workflows. Full training details can be found in the Appendix \ref{app:training_hyperparams}.

\subsection{Ablation: Effect of Multi-Scale Hierarchy}

To assess the contribution of the multiscale message-passing hierarchy, we conduct an ablation in which the diffusion
model architecture and training setup are kept identical, but the reduced-mesh pathway is removed. Specifically, we
replace the hierarchical MS--MGN denoiser with a single-scale baseline that retains only the original-mesh
message-passing network (the \texttt{o2o} branch, corresponding to step (1) in Figure~\ref{fig:architecture}), following the standard MGN architecture implemented with three processor layers. This ablation removes both the
global information transfer afforded by the \texttt{r2r} graph (step (3) in Figure~\ref{fig:architecture}) and the additional message-passing depth made computationally feasible at the reduced mesh resolution. As a result, the baseline performs all communication at the native mesh scale, which limits its ability to propagate long-range spatial dependencies efficiently and tends to emphasize local smoothing. By comparing the two models under matched diffusion settings and training conditions, we isolate the effect of hierarchical information flow on generation accuracy and spatial coherence.

Table~\ref{tab:flow_moments_compare} reports the flow moments aggregated by angle, namely the mean ($\mu$) and standard deviation ($\sigma$) of $U_x$, $U_y$ and $|\mathbf{u}|$, defined in Appendix~\ref{app:metrics}. These quantities characterize the large-scale directional tendencies and variability of the flow and therefore evaluate whether the generator recovers the bulk organization of the velocity field.

The multiscale model reproduces these flow moments markedly better than the single-scale model. The latter systematically underestimates both mean speeds and variability, reflecting excessive smoothing and loss of energetic content. In contrast, the multiscale architecture yields means and standard deviations close to the ground truth at both heights, indicating improved recovery of dominant flow directionality and intra-field variation.

Table~\ref{tab:metrics_compare} complements this analysis using the full suite of evaluation metrics defined in Appendix~\ref{app:metrics}. These include: relative error $L^2$ (magnitude accuracy), MAE (end-point velocity error), cosine similarity (directional alignment) and structure-function distance $\varepsilon_{S_2}$ (multiscale fluctuation reproduction). Lower values are better for all errors, whereas higher values indicate better alignment for cosine similarity.

Across all metrics, the multiscale hierarchy consistently improves accuracy at both altitudes. Relative errors in $L^2$ drop by
18--26\%, cosine similarity increases substantially, and the structure-function distance is reduced by a factor of 2--3 (e.g.\ 
$0.194\!\to\!0.067$ in $z{=}35$), indicating improved preservation of multiscale flow statistics. Together, these results show
that hierarchical message passing improves both global coherence and local structure recovery.

Although the single-scale model runs faster (Table~\ref{tab:metrics_compare}), the accuracy gap is significant. Using 20--40
denoising steps in the multiscale model yields a more favorable balance between sampling speed and physical fidelity.

\begin{table}
\caption{Flow moments at each $z$ (aggregated across angles). Quantities correspond to the mean ($\mu$) and standard deviation ($\sigma$) of the velocity components $U_x$, $U_y$, and speed $|\mathbf{u}|$, computed as defined in Appendix~\ref{app:metrics}. These statistics evaluate whether the generator reproduces large-scale directional trends and overall variability of the flow field.}

\label{tab:flow_moments_compare}
\centering
\small
\setlength{\tabcolsep}{4pt}
\renewcommand{\arraystretch}{1.1}
\begin{tabularx}{\linewidth}{l *{3}{c} *{3}{c}}
\toprule
& \multicolumn{3}{c}{$z=35$} & \multicolumn{3}{c}{$z=40$}\\
\cmidrule(lr){2-4}\cmidrule(lr){5-7}
\textbf{Quantity} 
& \textbf{GT} 
& \textbf{Multi-Scale} 
& \textbf{Single-Scale} 
& \textbf{GT} 
& \textbf{Multi-Scale} 
& \textbf{Single-Scale}\\
\midrule
$\mu_{U_x}$ 
& $-0.053 \pm 0.880$ & $-0.070 \pm 0.800$ & $-0.056 \pm 0.614$
& $-0.117 \pm 1.351$ & $-0.133 \pm 1.194$ & $-0.063 \pm 0.846$\\
$\mu_{U_y}$ 
& $0.132 \pm 0.844$ & $0.094 \pm 0.824$ & $0.122 \pm 0.616$
& $0.201 \pm 1.275$ & $0.164 \pm 1.172$ & $0.162 \pm 0.865$\\
$\mu_{|\mathbf{u}|}$ 
& $1.343 \pm 0.165$ & $1.240 \pm 0.119$ & $0.925 \pm 0.077$
& $1.888 \pm 0.210$ & $1.699 \pm 0.155$ & $1.266 \pm 0.128$\\
\addlinespace
$\sigma_{U_x}$ 
& $0.984 \pm 0.334$ & $0.915 \pm 0.352$ & $0.784 \pm 0.396$
& $1.026 \pm 0.366$ & $0.982 \pm 0.432$ & $0.927 \pm 0.456$\\
$\sigma_{U_y}$ 
& $0.977 \pm 0.276$ & $0.926 \pm 0.306$ & $0.821 \pm 0.354$
& $1.043 \pm 0.328$ & $0.994 \pm 0.315$ & $0.926 \pm 0.381$\\
$\sigma_{|\mathbf{u}|}$ 
& $1.242 \pm 0.101$ & $1.213 \pm 0.081$ & $1.153 \pm 0.079$
& $1.321 \pm 0.116$ & $1.311 \pm 0.134$ & $1.294 \pm 0.129$\\
\bottomrule
\end{tabularx}
\end{table}

\begin{table}
\caption{Generation performance metrics at each $z$ (aggregated across angles). Metrics follow the definitions in Appendix~\ref{app:metrics}: $\varepsilon_{\mathrm{rel},L^2}$ (relative $L^2$ error), MAE (end-point error), cosine similarity (directional agreement), and $\varepsilon_{S_2}$ (structure-function distance, assessing multiscale fluctuations). Lower values are better for all errors; higher is better for cosine similarity.}

\label{tab:metrics_compare}
\centering
\small
\renewcommand{\arraystretch}{1.1}
\resizebox{\linewidth}{!}{%
\begin{tabular}{l S[table-figures-integer=1,table-figures-decimal=3,table-figures-uncertainty=3]
                   S[table-figures-integer=1,table-figures-decimal=3,table-figures-uncertainty=3]
                   S[table-figures-integer=1,table-figures-decimal=3,table-figures-uncertainty=3]
                   S[table-figures-integer=1,table-figures-decimal=3,table-figures-uncertainty=3]}
\toprule
& \multicolumn{2}{c}{$z=35$} & \multicolumn{2}{c}{$z=40$}\\
\cmidrule(lr){2-3}\cmidrule(lr){4-5}
\textbf{Metric} & \textbf{Multi-Scale} & \textbf{Single-Scale} & \textbf{Multi-Scale} & \textbf{Single-Scale} \\
\midrule
$\varepsilon_{\mathrm{rel},L^{2}}(\mathbf{u})$   & 0.488 \pm 0.033 & 0.604 \pm 0.032 & 0.448 \pm 0.039 & 0.581 \pm 0.016 \\
$\varepsilon_{\mathrm{rel},L^{2}}(U_x)$          & 0.561 \pm 0.147 & 0.660 \pm 0.141 & 0.524 \pm 0.197 & 0.643 \pm 0.152 \\
$\varepsilon_{\mathrm{rel},L^{2}}(U_y)$          & 0.550 \pm 0.169 & 0.656 \pm 0.125 & 0.532 \pm 0.167 & 0.643 \pm 0.123 \\
$\varepsilon_{\mathrm{rel},L^{2}}(|\mathbf{u}|)$ & 0.399 \pm 0.031 & 0.526 \pm 0.040 & 0.376 \pm 0.031 & 0.513 \pm 0.014 \\
\addlinespace
\text{MAE}                                         & 0.685 \pm 0.090 & 0.822 \pm 0.125 & 0.767 \pm 0.099 & 1.025 \pm 0.108 \\
$\langle \cos\theta \rangle$                       & 0.560 \pm 0.063 & 0.487 \pm 0.034 & 0.711 \pm 0.029 & 0.609 \pm 0.044 \\
$\varepsilon_{S_2}$                                & 0.067 \pm 0.036 & 0.194 \pm 0.056 & 0.075 \pm 0.028 & 0.211 \pm 0.048 \\
\bottomrule
\end{tabular}
}
\end{table}

Considering the current number of layers, the multiscale message-passing hierarchy enables better long-range communication and spatial coherence during denoising, leading to superior reconstruction of both global flow trends and finer structural variations.

The results presented above highlight the strength and novelty of the proposed framework: to our knowledge, this is the first geometry-conditioned diffusion model capable of generating steady-state urban wind fields directly on CFD-grade 
unstructured meshes without requiring reference flow information at inference time. The model consistently reproduces spatial organization, multiscale statistics, distributional structure, and POD energy content across unseen geometries and wind angles. Importantly, these evaluations are performed on a particularly challenging dataset: the Bristol urban domain contains strong topographic variations, altitude-dependent building layouts, dense canyon structures, and 360 inflow angles per slice. This inherent geometric diversity provides a rigorous test of generalization, and the model performs robustly across all conditions, demonstrating its suitability as a practical surrogate for urban microclimate analysis. Such capabilities are directly relevant for real-time urban planning applications, early-stage design exploration, and computationally efficient digital-twin development.

While the approach is already effective, some aspects can be further improved in future work, for example by exploring deeper message-passing networks or expanding the dataset to additional cities and flow regimes. These represent natural extensions rather than fundamental shortcomings, and the present work provides a solid and technically innovative foundation upon which these future advances can build.

\section{Conclusions}
\label{sec:conclusion}

We introduce a geometry-conditioned diffusion framework that generates steady urban wind fields directly on unstructured meshes. In particular, the framework  requires only mesh information and wind direction as input. By coupling score-based diffusion with a hierarchical multi-resolution GNN denoiser, the model learns a probability distribution over plausible flows and produces diverse, uncertainty-aware samples without temporal rollouts, initialization fields, or modal bases. Trained in multiple altitude slices and wind angles, the approach generalizes to entirely unseen meshes, recovering physically salient structures such as wakes, shear layers, and recirculation zones.

Extensive experiments demonstrate that the generated fields reproduce not only pointwise velocities but also higher-level flow structures: spatial organization, vorticity patterns, angle-wise trends, distributional statistics, and multiscale energy content. A sampler study shows that accuracy saturates within a modest number of steps, yielding a favorable accuracy–cost balance suitable for interactive or real-time use. The ablation further highlights the impact of the hierarchical architecture, showing clear improvements over a single-scale GNN in both error metrics and multiscale coherence.

Beyond constituting an efficient surrogate for CFD acceleration, this framework enables a wide range of practical workflows in urban analysis. Fast sampling allows practitioners to explore many geometries and wind directions, evaluate microclimate comfort, perform safety and dispersion assessments, quantify uncertainty, and integrate flow generation into interactive decision-support tools. Furthermore, the generated wind fields can serve as initial conditions for downstream tasks such as the simulation of autonomous drone trajectories for environmental monitoring or pollution sensing, applications where rapid access to plausible wind fields is critical. Operating directly on unstructured meshes preserves geometric fidelity and makes the approach naturally portable across discretizations, supporting city-scale deployments with heterogeneous resolutions and domains.

This work represents, to the best of our knowledge, the first generative model operating on large, CFD-grade unstructured urban meshes, producing physically coherent flows conditioned solely on geometry and inflow direction. The combination of graph-based diffusion, mesh-aware conditioning, and multiscale message passing introduces a novel and effective paradigm for generating geometry-aware surrogates in complex built environments. These results suggest that diffusion on graphs can serve as a powerful foundation for future urban fluid-dynamics models.

We demonstrate that diffusion models on graphs offer a practical way to generate fast, geometry-aware urban wind surrogates. The iterative denoising process acts like a lightweight solver, while the graph representation keeps obstacles and mesh topology explicit. Together, these choices enable real-time microclimate checks, interactive design exploration, and continuous digital-twin updates.

This  work opens several promising research  directions. Incorporating additional modalities, such as pollutant concentration, temperature, or pedestrian-level turbulence indicators, would enable joint generative modeling of urban microclimate variables. Expanding the dataset to include multiple cities and a wider diversity of topographies could further enhance transferability. Finally, exploring deeper message-passing schemes or hybrid physical-neural architectures may yield even stronger global coherence while maintaining scalability.

\section*{Acknowledgments}

This work was supported by the Industrial PhD Program of the ‘Comunidad de Madrid’ under project reference IND2024/TIC-34540.

The authors acknowledge the MODELAIR project that has received funding from the European Union’s Horizon Europe research and innovation programme under the Marie Sklodowska-Curie grant agreement No. 101072559. The results of this publication reflect only the author’s views and do not necessarily reflect those of the European Union. The European Union can not be held responsible for them. S.L.C. acknowledges the grant PID2023-147790OB-I00 funded by MCIU/AEI/10.13039 /501100011033 /FEDER, UE. The authors gratefully acknowledge the Universidad Polit\'ecnica de Madrid (www.upm.es) for providing computing resources on the Magerit Supercomputer. PK acknowledges support by Harvard's Salata Institutre and Harvard's Data Science initiative.

\bibliographystyle{plainnat} 
\bibliography{references}

@inproceedings{
valencia2025learning,
title={Learning Distributions of Complex Fluid Simulations with Diffusion Graph Networks},
author={Mario Lino Valencia and Tobias Pfaff and Nils Thuerey},
booktitle={The Thirteenth International Conference on Learning Representations},
year={2025},
url={https://openreview.net/forum?id=uKZdlihDDn}
}

@misc{ho2020denoisingdiffusionprobabilisticmodels,
      title={Denoising Diffusion Probabilistic Models}, 
      author={Jonathan Ho and Ajay Jain and Pieter Abbeel},
      year={2020},
      eprint={2006.11239},
      archivePrefix={arXiv},
      primaryClass={cs.LG},
      url={https://arxiv.org/abs/2006.11239}, 
}

@inproceedings{Vishwasrao2024diffusion,
  author    = {A. Vishwasrao and S. Gutha and A. Patil and K. Wijk and B. McKeon and C. Gorle and H. Azizpour and R. Vinuesa},
  title     = {Diffusion models for optimal sensor placement and sparse reconstruction for simplified urban flows},
  booktitle = {Proceedings of the Summer Program, Center for Turbulence Research},
  year      = {2024},
}

@article{karras2022edm,
  title     = {Elucidating the Design Space of Diffusion-Based Generative Models},
  author    = {Karras, Tero and Aittala, Miika and Aila, Timo and Laine, Samuli},
  journal   = {Advances in Neural Information Processing Systems (NeurIPS)},
  year      = {2022},
  eprint    = {2206.00364},
  archivePrefix = {arXiv},
  primaryClass = {cs.CV}
}

@inproceedings{pfaff2021mgn,
  title     = {Learning Mesh-Based Simulation with Graph Networks},
  author    = {Pfaff, Tobias and Fortunato, Meire and Sanchez-Gonzalez, Alvaro and Battaglia, Peter W.},
  booktitle = {International Conference on Learning Representations (ICLR)},
  year      = {2021},
  eprint    = {2010.03409},
  archivePrefix = {arXiv},
  primaryClass = {cs.LG}
}

@article{fortunato2022msmgn,
  title     = {Multiscale MeshGraphNets},
  author    = {Fortunato, Meire and Pfaff, Tobias and Sanchez-Gonzalez, Alvaro and Wirnsberger, Peter and Battaglia, Peter},
  journal   = {arXiv preprint arXiv:2210.00612},
  year      = {2022},
  archivePrefix = {arXiv},
  primaryClass = {cs.LG}
}

@article{lam2023graphcast,
  title     = {GraphCast: Learning skillful medium-range global weather forecasting},
  author    = {Lam, Remi and Sanchez-Gonzalez, Alvaro and Doersch, Carl and others},
  journal   = {Science},
  volume    = {382},
  number    = {6677},
  pages     = {1416--1421},
  year      = {2023},
  doi       = {10.1126/science.adi2336}
}

@article{BenMoshe2023,
  title={Using machine learning to predict wind flow in urban areas},
  author={BenMoshe, Nir and Fattal, Eyal and Leitl, Bernd and Arav, Yehuda},
  journal={Atmosphere},
  volume={14},
  number={6},
  pages={990},
  year={2023},
  publisher={MDPI}
}

@article{Kastner2023GAN,
  title={A GAN-based surrogate model for instantaneous urban wind flow prediction},
  author={Kastner, Patrick and Dogan, Timur},
  journal={Building and Environment},
  volume={242},
  pages={110384},
  year={2023},
  publisher={Elsevier}
}

@article{Shao2023PIGNN,
  title={PIGNN-CFD: A physics-informed graph neural network for rapid predicting urban wind field defined on unstructured mesh},
  author={Shao, Xuqiang and Liu, Zhijian and Zhang, Siqi and Zhao, Zijia and Hu, Chenxing},
  journal={Building and Environment},
  volume={232},
  pages={110056},
  year={2023},
  publisher={Elsevier}
}

@article{Lu2023CanopyFlows,
  title={Using machine learning to predict urban canopy flows for land surface modeling},
  author={Lu, Yanle and Zhou, Xu-Hui and Xiao, Heng and Li, Qi},
  journal={Geophysical Research Letters},
  volume={50},
  number={1},
  pages={e2022GL102313},
  year={2023},
  publisher={Wiley Online Library}
}

@article{Yang2023FluidDiff,
  title={A denoising diffusion model for fluid field prediction},
  author={Yang, Gefan and Sommer, Stefan},
  journal={arXiv preprint arXiv:2301.11661},
  year={2023}
}

@article{Shu2023DiffusionPDE,
  title={A physics-informed diffusion model for high-fidelity flow field reconstruction},
  author={Shu, Dule and Li, Zijie and Farimani, Amir Barati},
  journal={Journal of Computational Physics},
  volume={478},
  pages={111972},
  year={2023},
  publisher={Elsevier}
}

@article{Whittaker2024TurbulenceDiffusion,
  title={Turbulence scaling from deep learning diffusion generative models},
  author={Whittaker, Tim and Janik, Romuald A and Oz, Yaron},
  journal={Journal of Computational Physics},
  volume={514},
  pages={113239},
  year={2024},
  publisher={Elsevier}
}

@inproceedings{Brandstetter2022MP-PDE,
  title={Message Passing Neural PDE Solvers},
  author={Brandstetter, Johannes and Worrall, Daniel E. and Welling, Max},
  booktitle={International Conference on Learning Representations (ICLR)},
  year={2022}
}

@article{Nguyen2025PhysiX,
  title={PhysiX: A Foundation Model for Physics Simulations},
  author={Nguyen, Tung and Koneru, Arsh and Li, Shufan and Grover, Aditya},
  journal={arXiv preprint arXiv:2506.17774},
  year={2025},
  url={https://arxiv.org/abs/2506.17774}
}

@article{Gao2025CMA_GuidedDiffusion,
  title={Generative learning of the solution of parametric Partial Differential Equations using guided diffusion models and virtual observations},
  author={Gao, Han and Kaltenbach, Sebastian and Koumoutsakos, Petros},
  journal={Computer Methods in Applied Mechanics and Engineering},
  volume={435},
  pages={117654},
  year={2025},
  doi={10.1016/j.cma.2024.117654}
}

@article{Nature2025,
  title={How cities can keep their cool},
  author={Eisenstein, Michael},
  journal={Nature},
  volume={648},
  year={2025},
  doi={doi: https://doi.org/10.1038/d41586-025-03926-4}
}

@article{Price2024GenCast,
  title={Probabilistic weather forecasting with machine learning},
  author={Price, Ilan and Sanchez-Gonzalez, Alvaro and Alet, Ferran and Andersson, Tom R. and El-Kadi, Andrew and Masters, Dominic and Ewalds, Timo and Stott, Jacklynn and Mohamed, Shakir and Battaglia, Peter and Lam, Remi and Willson, Matthew},
  journal={Nature},
  year={2024},
  doi={10.1038/s41586-024-08252-9}
}

@article{Wang2024CoNFiLD,
  title={Conditional neural field latent diffusion model for generating spatiotemporal turbulence},
  author={Du, Pan and Parikh, Meet Hemant and Fan, Xiantao and Liu, Xin-Yang and Wang, Jian-Xun},
  journal={Nature Communications},
  volume={15},
  number={1},
  pages={10416},
  year={2024},
  publisher={Nature Publishing Group UK London}
}

@article{Gao2025GraphLED,
  title={Learning Effective Dynamics across Spatio-Temporal Scales of Complex Flows},
  author={Gao, Han and Kaltenbach, Sebastian and Koumoutsakos, Petros},
  journal={arXiv preprint arXiv:2502.07990},
  year={2025},
  url={https://arxiv.org/abs/2502.07990}
}

@article{Vishwasrao2025DiffSPORT,
  title={Diff--SPORT: Diffusion-based Sensor Placement Optimization and Reconstruction of Turbulent flows in urban environments},
  author={Vishwasrao, Abhijeet and Gutha, Sai Bharath Chandra and Cremades, Andres and Wijk, Klas and Patil, Aakash and Gorle, Catherine and McKeon, Beverley J and Azizpour, Hossein and Vinuesa, Ricardo},
  journal={arXiv preprint arXiv:2506.00214},
  year={2025},
  url={https://arxiv.org/abs/2506.00214}
}

@article{bodnar2025foundation,
  title={A foundation model for the Earth system},
  author={Bodnar, Cristian and Bruinsma, Wessel P and Lucic, Ana and Stanley, Megan and Allen, Anna and Brandstetter, Johannes and Garvan, Patrick and Riechert, Maik and Weyn, Jonathan A and Dong, Haiyu and others},
  journal={Nature},
  pages={1--8},
  year={2025},
  publisher={Nature Publishing Group UK London}
}

@article{blocken2015computational,
  title={Computational Fluid Dynamics for urban physics: Importance, scales, possibilities, limitations and ten tips and tricks towards accurate and reliable simulations},
  author={Blocken, Bert},
  journal={Building and environment},
  volume={91},
  pages={219--245},
  year={2015},
  publisher={Elsevier}
}

@article{world2016ambient,
  title={Ambient air pollution: A global assessment of exposure and burden of disease},
  author={World Health Organization and others},
  journal={Clean Air Journal},
  volume={26},
  number={2},
  pages={6--6},
  year={2016}
}

@article{lateb2016use,
  title={On the use of numerical modelling for near-field pollutant dispersion in urban environments- A review},
  author={Lateb, M and Meroney, Robert N and Yataghene, M and Fellouah, H and Saleh, F and Boufadel, MC},
  journal={Environmental Pollution},
  volume={208},
  pages={271--283},
  year={2016},
  publisher={Elsevier}
}

@article{bonev2025fourcastnet,
  title={Fourcastnet 3: A geometric approach to probabilistic machine-learning weather forecasting at scale},
  author={Bonev, Boris and Kurth, Thorsten and Mahesh, Ankur and Bisson, Mauro and Kossaifi, Jean and Kashinath, Karthik and Anandkumar, Anima and Collins, William D and Pritchard, Michael S and Keller, Alexander},
  journal={arXiv preprint arXiv:2507.12144},
  year={2025}
}

@misc{barragan2025hybrinethybridneuralnetworkbasedframework,
      title={HybriNet-Hybrid Neural Network-based framework for Multi-Parametric Database Generation, Enhancement, and Forecasting}, 
      author={Guillermo Barragán and Ashton Hetherington and Arindam Sengupta and Rodrigo Abadía-Heredia and Jesús Garicano-Mena and Soledad Le Clainche},
      year={2025},
      eprint={2510.25625},
      archivePrefix={arXiv},
      primaryClass={physics.flu-dyn},
      url={https://arxiv.org/abs/2510.25625}, 
}

@article{abadia2022predictive,
  title={A predictive hybrid reduced order model based on proper orthogonal decomposition combined with deep learning architectures},
  author={Abad{\'\i}a-Heredia, R and L{\'o}pez-Mart{\'\i}n, M and Carro, B and Arribas, JI and P{\'e}rez, JM and Le Clainche, S},
  journal={Expert Systems with Applications},
  volume={187},
  pages={115910},
  year={2022},
  publisher={Elsevier}
}

@misc{zou2025generativeartificialintelligencehybrid,
      title={Generative artificial intelligence and hybrid models to accelerate LES in reactive flows: Application to hydrogen/methane combustion}, 
      author={Xiangrui Zou and Rodrigo Abadia-Heredia and Laura Saavedra and Alessandro Parente and Rui Xue and Soledad Le Clainche},
      year={2025},
      eprint={2507.08426},
      archivePrefix={arXiv},
      primaryClass={physics.flu-dyn},
      url={https://arxiv.org/abs/2507.08426}, 
}

@article{wang2004image,
  title={Image quality assessment: from error visibility to structural similarity},
  author={Wang, Zhou and Bovik, Alan C and Sheikh, Hamid R and Simoncelli, Eero P},
  journal={IEEE transactions on image processing},
  volume={13},
  number={4},
  pages={600--612},
  year={2004},
  publisher={IEEE}
}

@article{chen2021adaspeech,
  title={Adaspeech: Adaptive text to speech for custom voice},
  author={Chen, Mingjian and Tan, Xu and Li, Bohan and Liu, Yanqing and Qin, Tao and Zhao, Sheng and Liu, Tie-Yan},
  journal={arXiv preprint arXiv:2103.00993},
  year={2021}
}

@article{molinaro2024generative,
  title={Generative ai for fast and accurate statistical computation of fluids},
  author={Molinaro, Roberto and Lanthaler, Samuel and Raoni{\'c}, Bogdan and Rohner, Tobias and Armegioiu, Victor and Simonis, Stephan and Grund, Dana and Ramic, Yannick and Wan, Zhong Yi and Sha, Fei and others},
  journal={arXiv preprint arXiv:2409.18359},
  year={2024}
}

@book{silverman2018density,
  title={Density estimation for statistics and data analysis},
  author={Silverman, Bernard W},
  year={2018},
  publisher={Routledge}
}

@article{panaretos2019statistical,
  title={Statistical aspects of Wasserstein distances},
  author={Panaretos, Victor M and Zemel, Yoav},
  journal={Annual review of statistics and its application},
  volume={6},
  number={1},
  pages={405--431},
  year={2019},
  publisher={Annual Reviews}
}

\clearpage
\appendix

\section{Additional Model and Training Details}
\subsection{Graph neural network message passing and conditioning}
\label{app:gnn_details}

This appendix summarizes the message-passing and conditioning equations used in the denoiser described in Section~\ref{sec:denoiser_architecture}. The notation follows that of GraphCast~\cite{lam2023graphcast} and MultiScale MeshGraphNets~\cite{fortunato2022msmgn}, simplified to two mesh levels: the original mesh (index $o$) and the reduced mesh (index $r$). Each of the four subnetworks (\texttt{o2o}, \texttt{o2r}, \texttt{r2r}, \texttt{r2o}) follows an \textit{encode--process--decode} structure similar to MeshGraphNets~\cite{pfaff2021mgn}.

Each mesh level $l \in \{o, r\}$ defines a graph $\mathcal{G}^l = (\mathcal{V}^l, \mathcal{E}^l)$, where $\mathcal{V}^l$ is the set of nodes and $\mathcal{E}^l$ the set of directed edges. Each node $i \in \mathcal{V}^l$ carries a feature vector $\mathbf{h}_i^l \in \mathbb{R}^{d_l}$, and each edge $(i,j) \in \mathcal{E}^l$ has an associated feature vector $\mathbf{e}_{ij}^l$. For cross-level mappings (\texttt{o2r} or \texttt{r2o}), we additionally consider heterogeneous graphs $\mathcal{G}^{lr} = (\mathcal{V}^l, \mathcal{V}^r, \mathcal{E}^{lr})$ with edges $i \to j$ and features $\mathbf{e}_{ij}^{lr}$ constructed from relative node positions and node-type identifiers. All edge features are based on $(dx, dy)$, the distance $d = \lVert \mathbf{x}_j - \mathbf{x}_i\rVert$, and the directional projections $p_{\mathrm{par}}$ and $p_{\mathrm{perp}}$ defined with respect to the global wind direction $\Phi$, as defined in Section~\ref{sec:denoiser_architecture}.

Before message passing, the raw node inputs (defined in Section~\ref{sec:denoiser_architecture}) and the raw edge inputs $\mathbf{f}_{ij}$ are mapped into latent space through encoder MLPs:
\begin{align}
\mathbf{h}_i^{(0)} &= \phi_{\text{enc}}^n(\mathbf{x}_i), \\
\mathbf{e}_{ij}^{(0)} &= \phi_{\text{enc}}^e(\mathbf{f}_{ij}),
\end{align}
where $\mathbf{h}_i^{(0)}$ and $\mathbf{e}_{ij}^{(0)}$ serve as the initial node and edge embeddings for message passing.

Each subnetwork performs $K$ message-passing iterations. At iteration $k$, node embeddings $\mathbf{h}_i^{(k)}$ and edge embeddings $\mathbf{e}_{ij}^{(k)}$ are updated through learned message functions. For each edge $(i,j)$, we compute a message
\begin{equation}
\mathbf{m}_{ij} = \phi_e\!\left(\mathbf{h}_i^{(k)},\, \mathbf{h}_j^{(k)},\, \mathbf{e}_{ij}^{(k)},\, \mathbf{g}\right),
\end{equation}
where $\mathbf{g}$ is a global conditioning vector formed by encoding the diffusion noise level $\sigma$ and the wind direction $\Phi$. Incoming messages are aggregated per node using a permutation-invariant operator,
\begin{equation}
\mathbf{m}_i = \sum_{j:(i,j)\in\mathcal{E}} \rho_e(\mathbf{m}_{ij}),
\end{equation}
and the node state is updated as
\begin{equation}
\mathbf{h}_i^{(k+1)} = \phi_h\!\left(\mathbf{h}_i^{(k)},\, \mathbf{m}_i,\, \mathbf{g}\right).
\end{equation}
Here, $\phi_e$ and $\phi_h$ denote MLPs applied at the edge and node levels, respectively, and $\rho_e(\cdot)$ denotes message aggregation by summation.

The global conditioning vector $g$ modulates the normalization layers in all processors \cite{chen2021adaspeech}, following the norm-conditioning scheme of GenCast~\cite{Price2024GenCast}. It is implemented through an affine normalization transform:
\begin{equation}
\text{LayerNorm}(\mathbf{h}_i; \mathbf{g})
=
\gamma(\mathbf{g}) \frac{\mathbf{h}_i - \mu(\mathbf{h})}{\sigma(\mathbf{h})} + \beta(\mathbf{g}),
\end{equation}
where $\gamma(\mathbf{g})$ and $\beta(\mathbf{g})$ are scale and bias terms predicted by an MLP applied to $\mathbf{g}$. This mechanism injects the effect of $\sigma$ and $\Phi$ consistently across layers and scales.

After $K$ message-passing iterations, each node’s latent representation can be decoded back to the target dimensionality:
\begin{equation}
\widehat{\mathbf{y}}_i = \phi_{\text{dec}}(\mathbf{h}_i^{(K)}),
\end{equation}
where $\phi_{\text{dec}}$ is an MLP with residual connections. In practice, this operation is only performed in the final \texttt{r2o} network, to produce the denoised velocity prediction on the original mesh.

The projection from original to reduced mesh (downsample) and the reverse mapping (upsample) are performed on heterogeneous graphs:
\begin{align}
\mathbf{h}_j^{r'} &= 
\phi_{r2r}\!\left(
\mathbf{h}_j^r, \sum_{i:(i,j)\in\mathcal{E}^{o2r}} \rho(\phi_e^{o2r}(\mathbf{h}_i^o, \mathbf{e}_{ij}^{o2r}))
\right), \\
\mathbf{h}_i^{o'} &= 
\phi_{o2o}\!\left(
\mathbf{h}_i^o, \sum_{j:(j,i)\in\mathcal{E}^{r2o}} \rho(\phi_e^{r2o}(\mathbf{h}_j^r, \mathbf{e}_{ji}^{r2o}))
\right).
\end{align}
This bidirectional exchange allows coarse-scale information from the reduced mesh to influence fine-scale updates on the original mesh, improving spatial coherence and enabling efficient long-range communication.

After the last upsample step, the node embeddings on the original mesh are projected to the output channels using a linear layer applied independently at each node,
\begin{equation}
\widehat{\mathbf{y}}_i = W_{\text{out}}\, \mathbf{h}_i^{o'} + b_{\text{out}}, \qquad i \in \mathcal{V}^o,
\end{equation}
which yields the predicted flow field $\widehat{\mathbf{Y}} = \{\widehat{\mathbf{y}}_i\}_{i\in\mathcal{V}^o}$ over all nodes. The output is then combined with the input according to the EDM preconditioning coefficients $(c_{\text{skip}}, c_{\text{out}})$ introduced in Section~\ref{sec:edm_background}.

Each of the four subnetworks thus contains a full encode--process--decode pipeline, with shared design principles but distinct graph connectivity. Together, they implement a hierarchical GNN system that enables both local fine-scale corrections and long-range contextual propagation within a single denoising iteration.

\subsection{Training and Hyperparameters}
\label{app:training_hyperparams}

Training follows the denoising score-matching formulation of EDM~\cite{karras2022edm}, where the model learns to reverse the noise corruption process across a continuous range of scales $\sigma \in [\sigma_{\min}, \sigma_{\max}]$. The training noise levels follow a log-log schedule:

\begin{equation}
\sigma(\rho, u) = \left( \sigma_{\max}^{1/\rho} + u \cdot (\sigma_{\min}^{1/\rho} - \sigma_{\max}^{1/\rho}) \right)^{\rho},
\end{equation}

where $u \sim \mathcal{U}(0,1)$ and $\rho = 7.0$. We use $\sigma_{\min} = 0.02 \cdot \frac{0.2757}{0.5}$ and $\sigma_{\max} = 88.0 \cdot \frac{0.2757}{0.5}$ as preconditioning bounds, scaled to match the empirical velocity scale of the dataset ($\sigma_{data}$). 

The model is trained using the AdamW optimizer with cosine learning rate decay. The base learning rate is $1\mathrm{e}{-4}$ with $\beta_1=0.9$, $\beta_2=0.999$, weight decay of $1\mathrm{e}{-2}$, and gradient clipping at a global norm of 5.0. We train for 200{,}000 steps with a batch size of 2 per GPU, distributed across 4 NVIDIA A100 GPUs using data parallelism.

All node and edge embeddings throughout the architecture use a hidden size of 64. The global conditioning vector--formed by concatenating the noise level $\sigma$ and wind direction $\Phi$--is also projected onto a 64-dimensional embedding. All MLPs and message-passing layers operate on this shared dimensionality.

For inference, we employ stochastic EDM sampling with $N{=}20$ steps. The sampler uses the same $\rho{=}7.0$ noise schedule, with $\sigma_{\min} = 0.02 \cdot \frac{0.2757}{0.5}$, $\sigma_{\max} = 80.0 \cdot \frac{0.2757}{0.5}$, and noise parameters $s_{\text{churn}}=2.5$, $s_{\min}=0.75$, $s_{\max}=\infty$, and $s_{\text{noise}}=1.05$. The conditioning vector allows the model to dynamically adapt to each diffusion step.

For all hyperparameters, we follow the same configuration as in \cite{Price2024GenCast} for both training and sampling. The only exception is the choice of $\sigma_{\text{data}}$, which is set directly from our dataset statistics. Since we do not apply standardization by mean and standard deviation, and instead scale the velocity components by their maximum values, the original $\sigma_{\text{data}}$ value is not appropriate in our setting. We found that using the dataset-specific $\sigma_{\text{data}}$ yields improved training stability and reconstruction quality.

\subsection{Evaluation Metrics}
\label{app:metrics}

We evaluate the predicted nodewise velocity field $\mathbf{u}^{\mathrm{pred}}_i=(U_{x,i},U_{y,i})$ against the ground-truth
$\mathbf{u}^{\mathrm{true}}_i$ using mesh-agnostic metrics computed directly on the original unstructured mesh. For clarity,
we denote the velocity magnitude at each node as $|\mathbf{u}_i|=\sqrt{U_{x,i}^2+U_{y,i}^2}$. All metrics are computed per
wind direction and then aggregated across directions as mean~$\pm$~standard deviation.

For any scalar or vector quantity $q\in\{\mathbf{u},U_x,U_y,|\mathbf{u}|\}$, we report the normalized root-mean-square
difference
\[
\varepsilon_{\mathrm{rel},L^{2}}(q)=
\frac{\sqrt{\tfrac{1}{N}\sum_{i=1}^{N}\|q^{\mathrm{pred}}_i - q^{\mathrm{true}}_i\|^2}}
     {\sqrt{\tfrac{1}{N}\sum_{i=1}^{N}\|q^{\mathrm{true}}_i\|^2}}.
\]
This error measures overall magnitude disagreement while accounting for the scale of the flow field.

The mean absolute error (MAE) between predicted and true velocities is
\[
\mathrm{MAE}=\frac{1}{N}\sum_{i=1}^{N}\|\mathbf{u}^{\mathrm{pred}}_i - \mathbf{u}^{\mathrm{true}}_i\|.
\]
This corresponds to the standard End-Point Error (EPE) commonly used in optical-flow evaluation.

To assess directional alignment independently of magnitude, we compute the Cosine similarity
\[
\langle \cos\theta \rangle =
\frac{1}{N}\sum_{i=1}^{N}
\frac{\mathbf{u}^{\mathrm{pred}}_i\cdot\mathbf{u}^{\mathrm{true}}_i}
     {\|\mathbf{u}^{\mathrm{pred}}_i\|\;\|\mathbf{u}^{\mathrm{true}}_i\|+\varepsilon}.
\]
The small constant $\varepsilon$ prevents division by zero when one of the vectors has near-zero speed; this metric is
also known as the Modal Assurance Criterion (MAC) when used to assess directional agreement.

We interpolate $|\mathbf{u}|$ onto a regular grid and compute SSIM~\cite{wang2004image} to evaluate structural
coherence of spatial patterns. Values lie in $[0,1]$, with 1 indicating identical structure.

To quantify multiscale velocity fluctuations, we compute the second-order structure function of velocity magnitude,
defined for spatial separation distance $r$ as
\[
S_2(r)=\big\langle(|\mathbf{u}|(\mathbf{x+r})-|\mathbf{u}|(\mathbf{x}))^2\big\rangle.
\]
Here, separations $r$ correspond to Euclidean distances between mesh node pairs, binned into discrete separation ranges.
We denote by $S_{2}^{\mathrm{pred}}(r_k)$ and $S_{2}^{\mathrm{true}}(r_k)$ the predicted and ground-truth values for bin $k$, and $w_k$
are normalized bin weights proportional to the number of node pairs contributing. The relative curve distance is
\[
\varepsilon_{S_2}=\frac{\sqrt{\sum_k w_k\left(S_2^{\mathrm{pred}}(r_k)-S_2^{\mathrm{true}}(r_k)\right)^2}}
{\sqrt{\sum_k w_k\left(S_2^{\mathrm{true}}(r_k)\right)^2}+\varepsilon}.
\]
Lower values indicate better reproduction of the flow's spatial fluctuation spectrum.

For each angle, we compute the nodewise mean $\mu$ and standard deviation $\sigma$ of $U_x$, $U_y$, and $|\mathbf{u}|$
across the full mesh. These summarize bulk directional trends and variability and are then aggregated across angles
(see Tables~\ref{tab:flow_moments_compare}). In addition, probability density functions (PDFs) of these quantities are
estimated using Kernel Density Estimation (KDE) to assess distributional fidelity.

We report the mean wall-clock time required to generate a full slice at a given altitude $z$, providing a practical metric
for computational efficiency.

\section{Dataset Details}
\label{app:dataset_details}

\begin{figure}[t]
  \centering
  \begin{subfigure}[t]{0.32\linewidth}
    \centering
    \includegraphics[width=\linewidth]{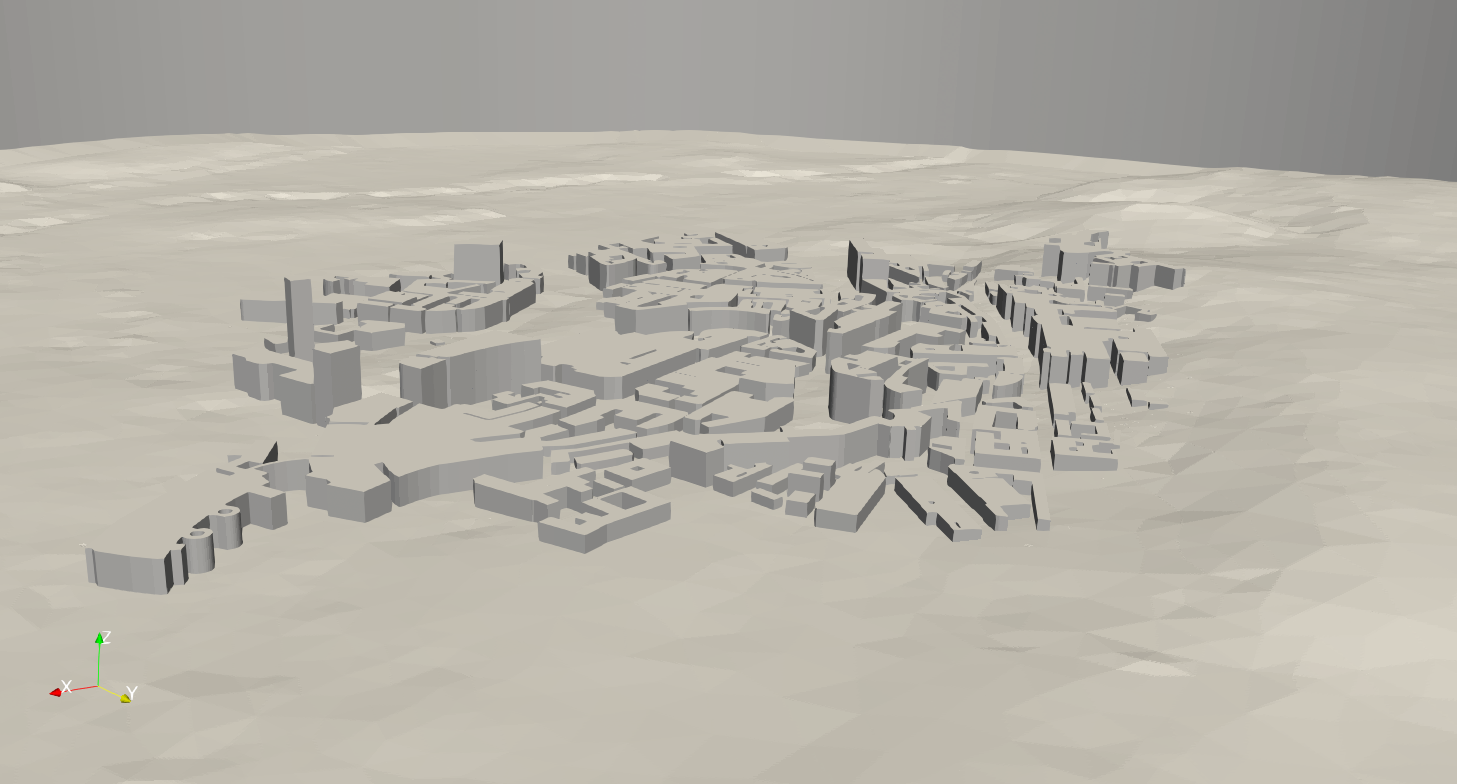}
    \caption{}
  \end{subfigure}
  \hfill
  \begin{subfigure}[t]{0.32\linewidth}
    \centering
    \includegraphics[width=\linewidth]{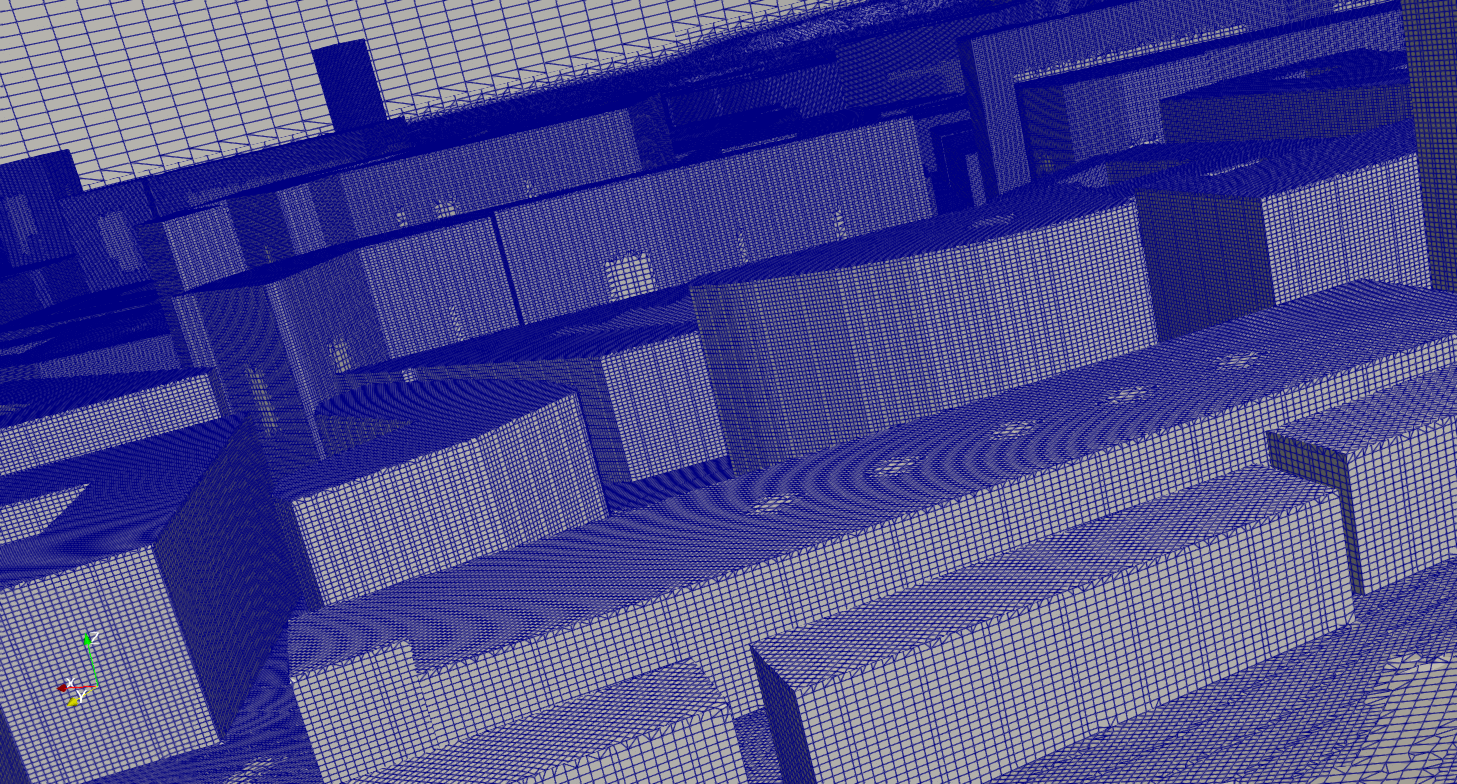}
    \caption{}
  \end{subfigure}
  \hfill
  \begin{subfigure}[t]{0.32\linewidth}
    \centering
    \includegraphics[width=\linewidth]{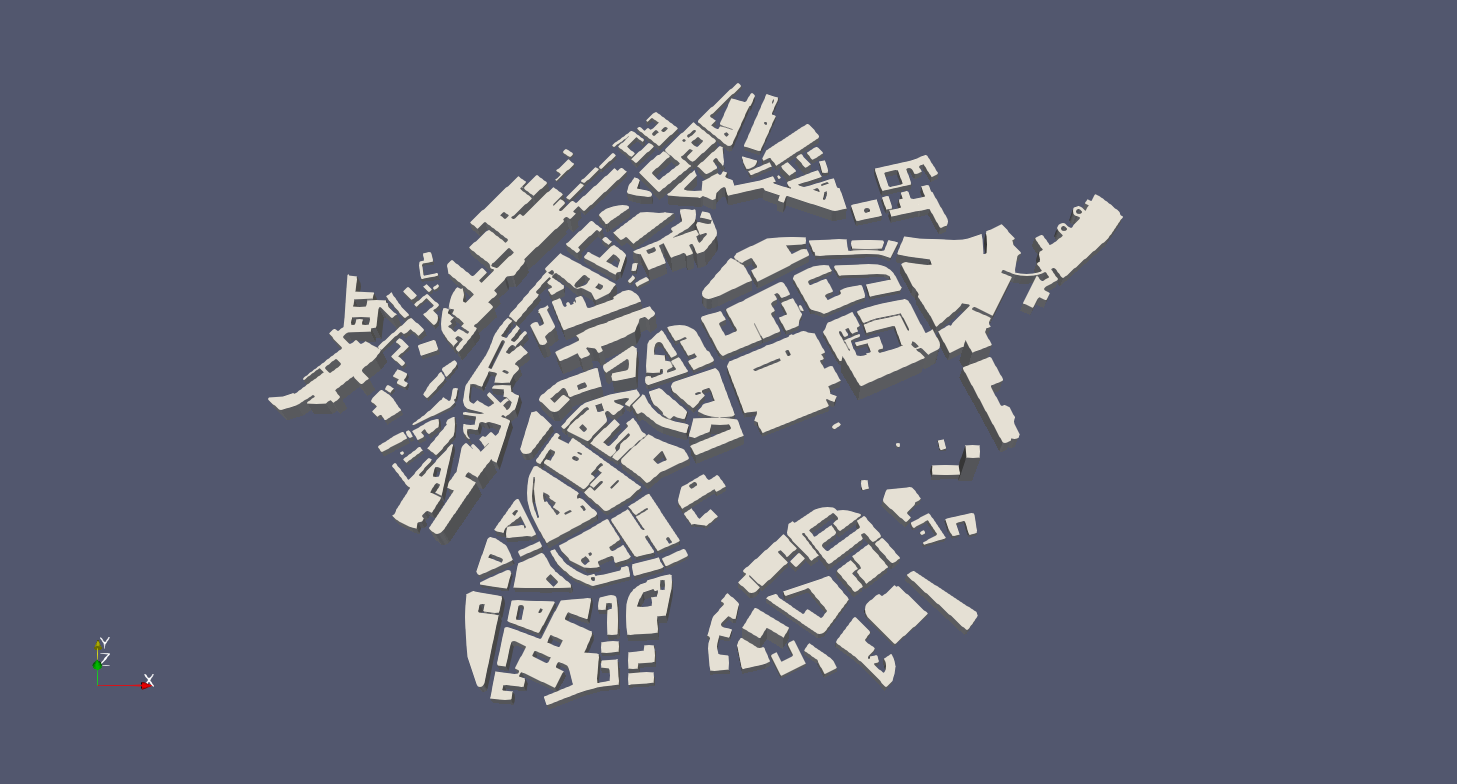}
    \caption{}
  \end{subfigure}
  \caption{3D CFD domain used to generate the dataset. The geometry includes a realistic terrain profile and detailed building blocks, enclosed within a cylindrical control volume. (a) Urban domain and terrain. (b) Volumetric mesh close-up. (c) Upside-down view of the 3D city geometry.}
  \label{fig:3d_meshes}
\end{figure}

The base dataset corresponds to a high-resolution CFD domain representing an urban area with realistic topography and detailed building geometry. The 3D control volume is cylindrical and discretized into approximately 40 million unstructured cells. Figure~\ref{fig:3d_meshes} illustrates the full 3D setup and mesh characteristics at different zoom levels. More information about the database can be found at \href{https://modelair.eu/}{modelair.eu}.

To construct a 2D dataset suitable for our graph-based diffusion model, we extract horizontal slices from the 3D domain at six altitudes ($z=\{15,20,28,35,40,45\}$\,m). Each slice produces a different 2D mesh, capturing unique building cross-sections and terrain variations. The domain is cropped to $[-1000,1000]\!\times\![-1000,1000]$\,m and decimated to maintain comparable node counts across slices.

Figure \ref{fig:training_and_test_data} shows training and test slices and how they change in topology, as well as the differences produced in the flow by some examples of the direction of the inflow wind.

Each 2D mesh includes node-type annotations distinguishing between fluid, wall, and boundary nodes. This information is used as categorical conditioning during training.  


For each slice, the dataset contains 360 stationary flow fields, each corresponding to a distinct wind direction.  
The fields are represented as $(U_x, U_y)$ velocity components per node.  
We use four slices for training and two for testing, ensuring that test geometries are unseen during training.

For every slice, both the original and reduced meshes are retained, yielding the four edge sets used by the hierarchical denoiser:  
\texttt{o2o}, \texttt{o2r}, \texttt{r2r}, and \texttt{r2o} (Figure \ref{fig:architecture}).  
These graphs maintain a consistent structure across all slices, ensuring comparable learning complexity.

Table~\ref{tab:mesh_statistics} summarizes the number of nodes and edges in both the original and reduced meshes for each altitude slice.  
These values indicate the scale of the graph structures used during training and testing, highlighting the balance between geometric richness and computational feasibility.

\begin{table}[t]
  \centering
  \caption{Number of nodes and edges for each slice and for each graph of the hierarchical architecture. Each slice corresponds to a 2D section of the original 3D CFD domain. The number of nodes and edges defines the scale of the graph input for the diffusion denoiser.}
  \label{tab:mesh_statistics}
  \begin{tabular}{lrrrrrr}
  \toprule
  Slice & Orig.\ nodes & Red.\ nodes & o2o edges & o2r edges & r2r edges & r2o edges \\
  \midrule
  $z_{15}$ & 302{.}011 & 62{.}929 & 1{.}686{.}618 & 239{.}082 & 298{.}054 & 239{.}082 \\
  $z_{20}$ & 302{.}043 & 54{.}305 & 1{.}674{.}914 & 247{.}738 & 278{.}770 & 247{.}738 \\
  $z_{28}$ & 301{.}532 & 56{.}889 & 1{.}689{.}602 & 244{.}643 & 286{.}802 & 244{.}643 \\
  $z_{35}$ & 310{.}655 & 56{.}319 & 1{.}777{.}546 & 254{.}336 & 291{.}246 & 254{.}336 \\
  $z_{40}$ & 305{.}735 & 58{.}222 & 1{.}783{.}482 & 247{.}513 & 300{.}044 & 247{.}513 \\
  $z_{45}$ & 300{.}780 & 55{.}557 & 1{.}766{.}300 & 245{.}223 & 296{.}916 & 245{.}223 \\
  \bottomrule
  \end{tabular}
\end{table}

\section{Additional Results}
\subsubsection{Additional Field Comparisons}
\begin{figure*}[ht]
  \centering
  \includegraphics[width=0.95\linewidth]{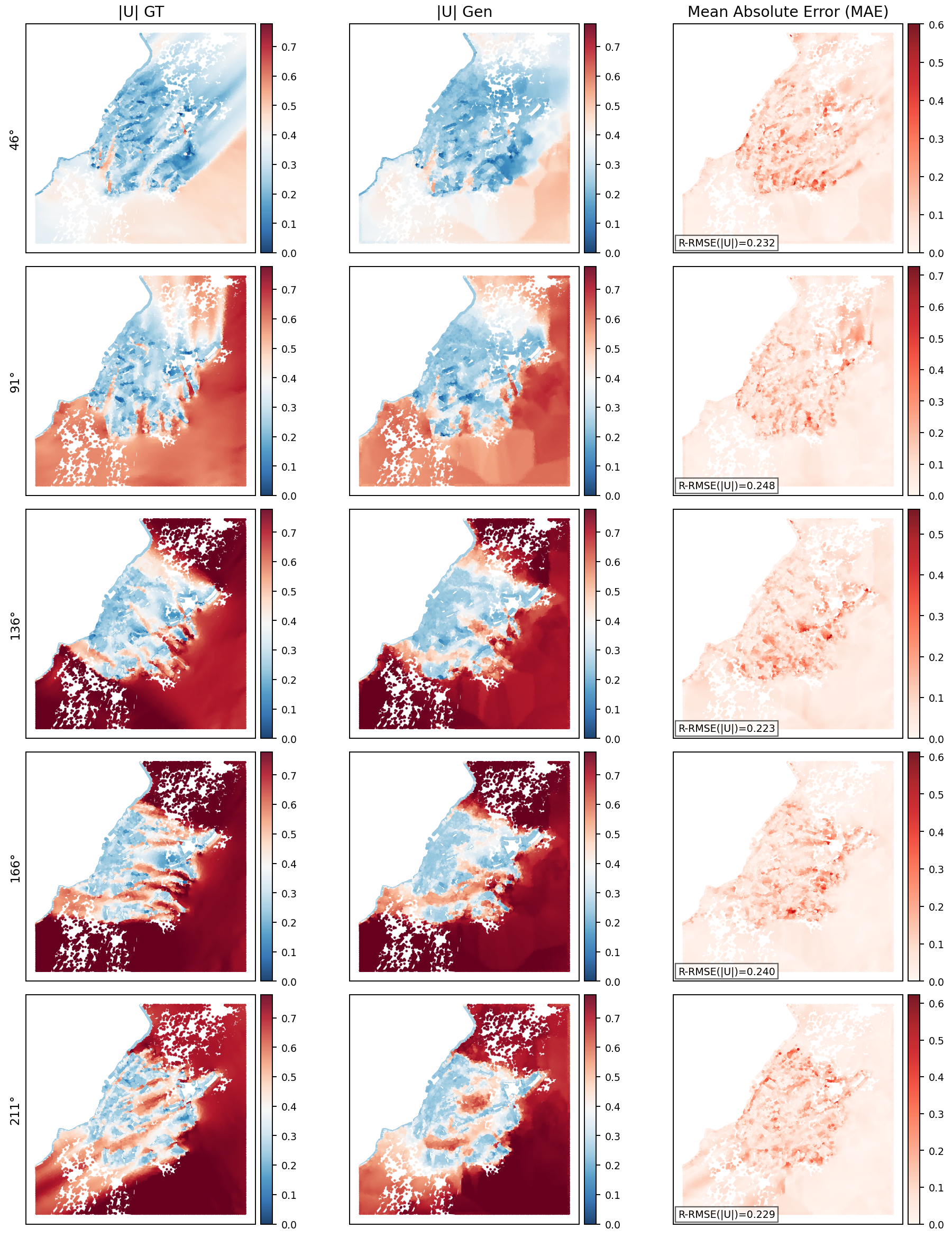}
  \caption{Comparison of wind magnitude $|\mathbf{u}|$ at $z{=}35$\,m across directions. Left to right: ground truth, generated output, and per-node mean absolute error (MAE). Relative RMSE is also shown per direction.}
  \label{fig:umag_appendix_z35}
\end{figure*}

\begin{figure*}[ht]
  \centering
  \includegraphics[width=0.95\linewidth]{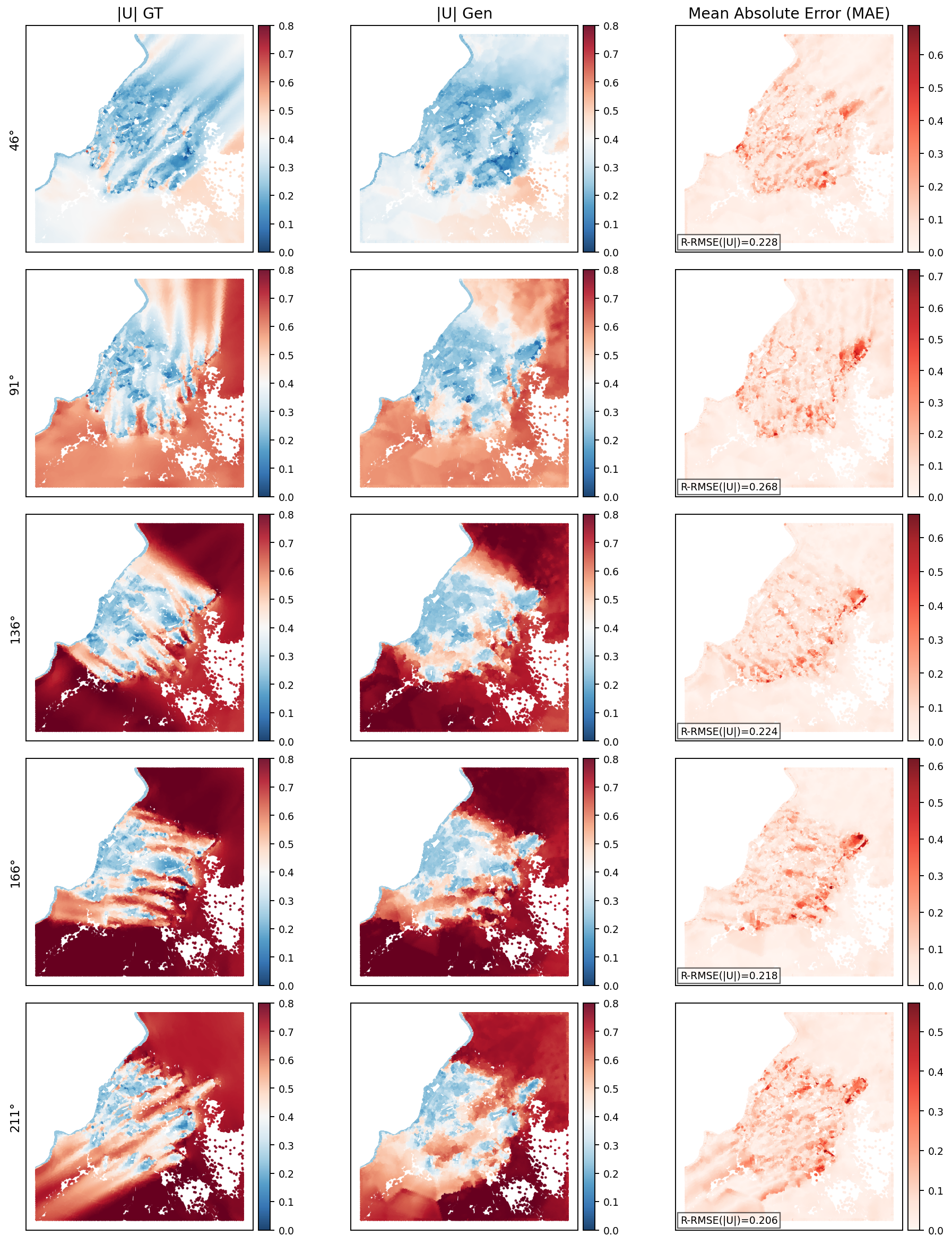}
  \caption{Comparison of wind magnitude $|\mathbf{u}|$ at $z{=}40$\,m. Similar trends to $z{=}35$\,m with mild degradation due to increased complexity and reduced geometric guidance at higher altitudes.}
  \label{fig:umag_appendix_z40}
\end{figure*}

Figure~\ref{fig:umag_appendix_z35} and Figure~\ref{fig:umag_appendix_z40} provide extended comparisons between ground truth and generated wind magnitude $|\mathbf{u}|$ fields at $z{=}35$\,m and $z{=}40$\,m, respectively. The right column shows per-node MAE and Mean Relative RMSE scores, revealing the spatial distribution of prediction errors.

\subsubsection{Zoomed Urban Flow Patterns}

\begin{figure*}[ht]
  \centering
  \includegraphics[width=0.9\linewidth]{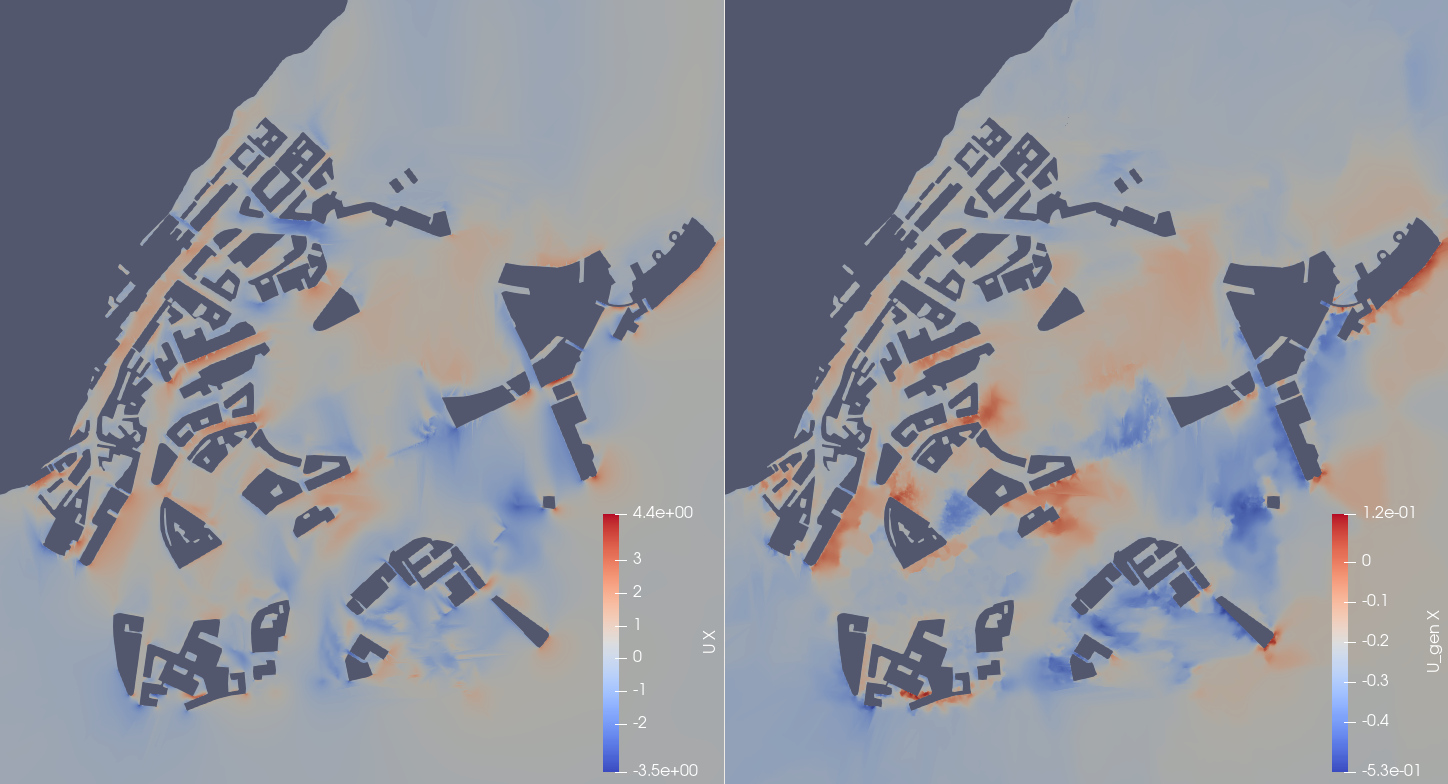}
  \caption{Zoomed $U_x$ field at $z{=}35$\,m, angle $91^\circ$. Ground-truth (left) vs.\ generated (right). The model captures canyon blockage and flow inversion near coastal buildings.}
  \label{fig:zoom_ux_91}
\end{figure*}

\begin{figure*}[ht]
  \centering
  \includegraphics[width=0.9\linewidth]{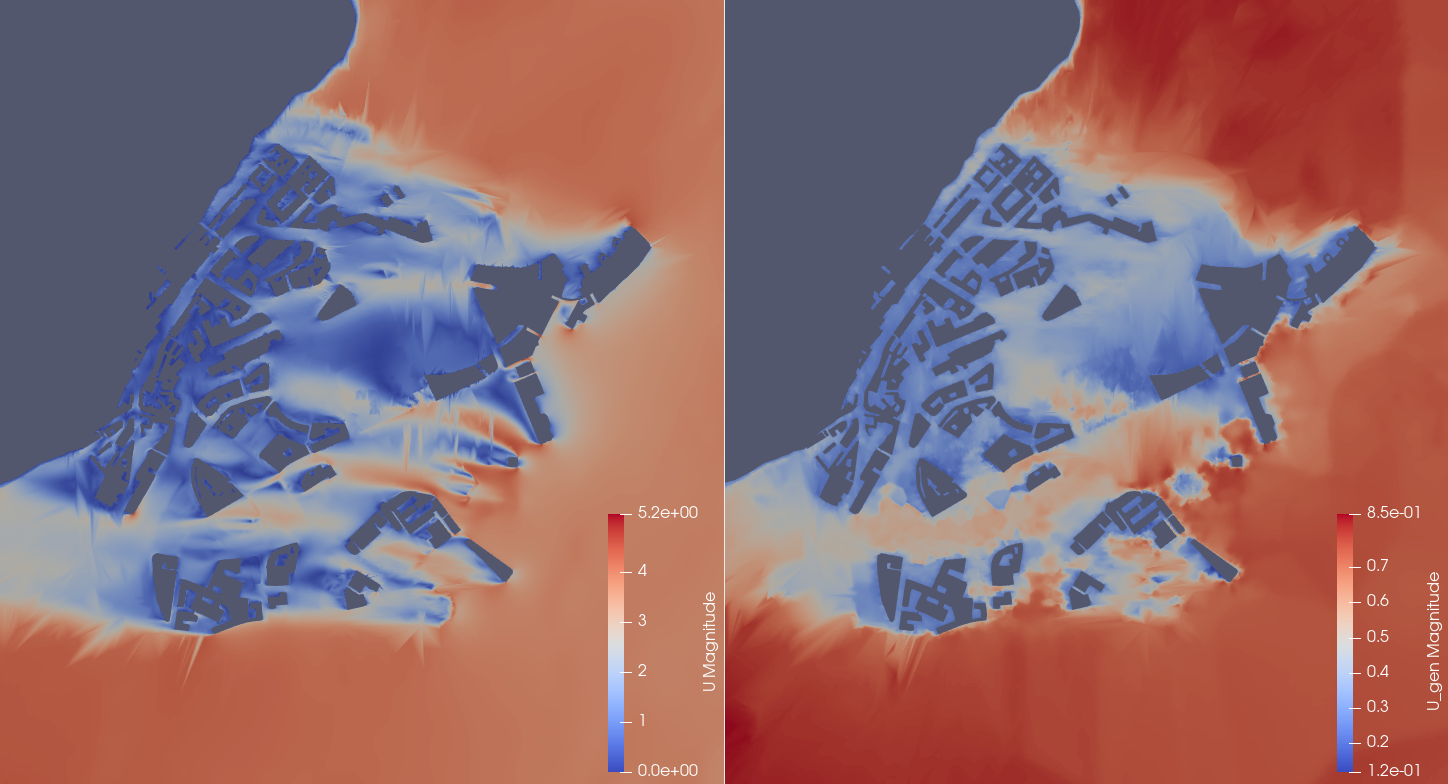}
  \caption{Zoomed $|\mathbf{u}|$ magnitude at $z{=}35$\,m, angle $166^\circ$. Ground-truth (left) vs.\ generated (right). Strong flow corridors and wake shadows behind buildings are preserved.}
  \label{fig:zoom_umag_166}
\end{figure*}

\begin{figure*}[ht]
  \centering
  \includegraphics[width=0.9\linewidth]{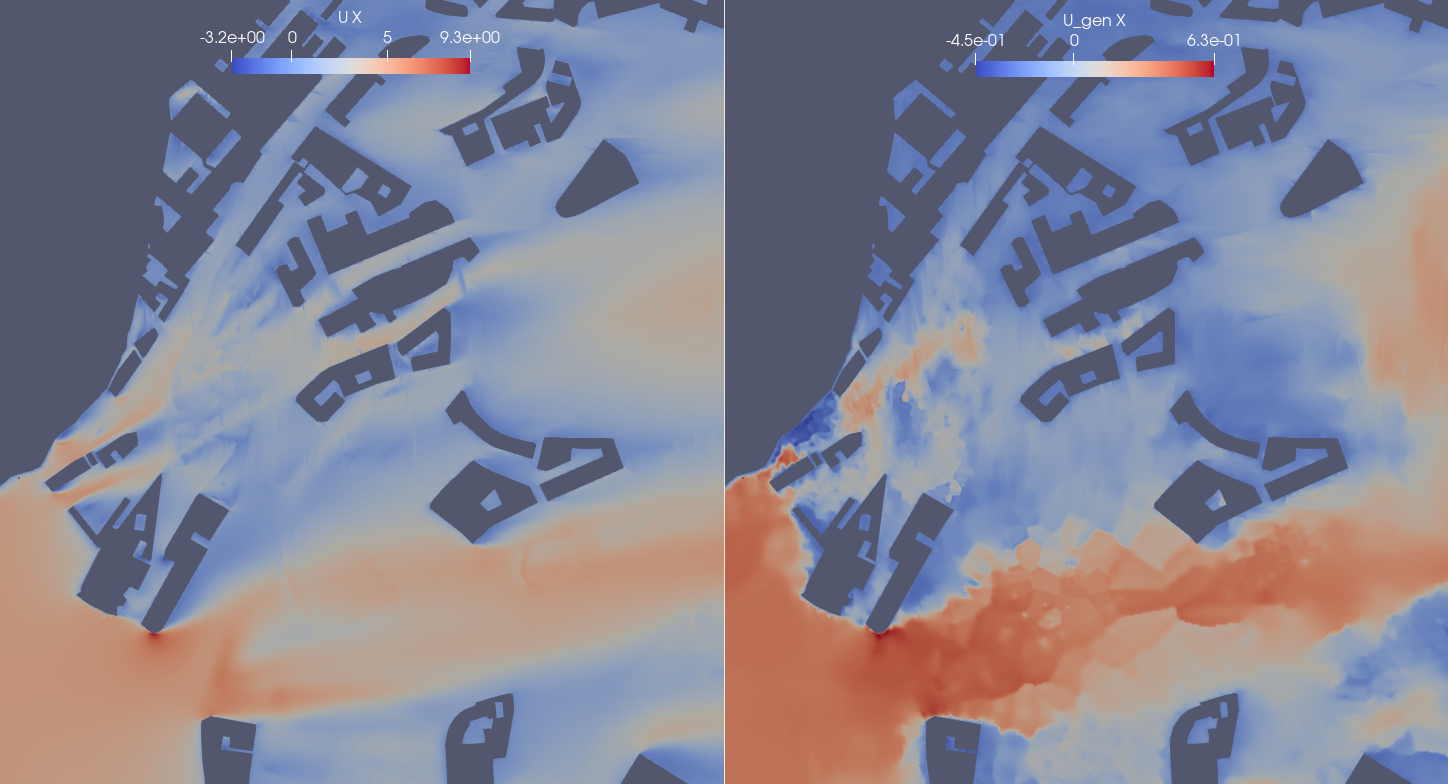}
  \caption{Zoomed full-slice field at $z{=}40$\,m, angle $1^\circ$. Ground-truth (left) vs.\ generated (right).}
  \label{fig:zoom_z40_angle1}
\end{figure*}

\begin{figure*}[ht]
  \centering
  \includegraphics[width=0.9\linewidth]{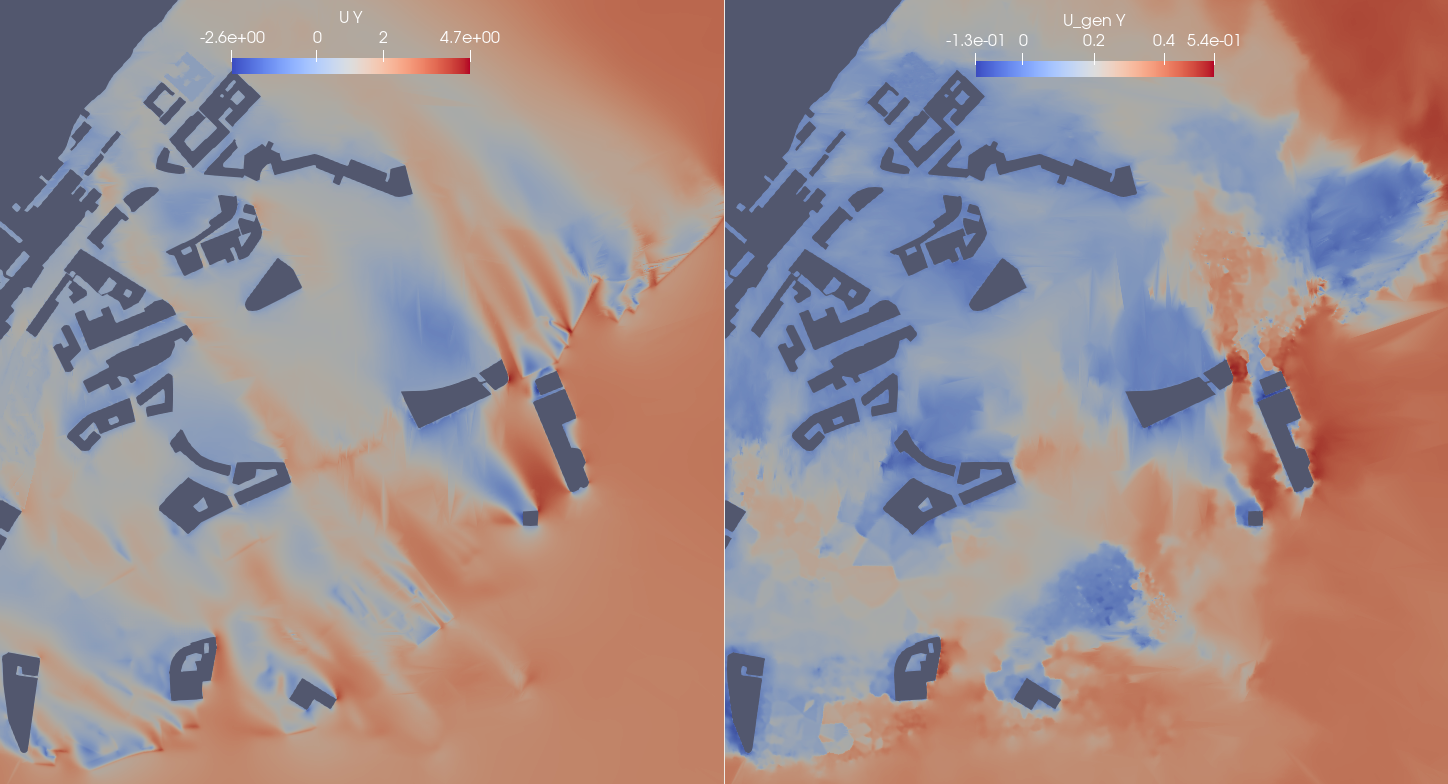}
  \caption{Zoomed $U_y$ field at $z{=}40$\,m, angle $136^\circ$. Ground-truth (left) vs.\ generated (right). Fine-scale vortical structures are captured.}
  \label{fig:zoom_z40_uy}
\end{figure*}

In Figures \ref{fig:zoom_ux_91}, \ref{fig:zoom_umag_166}, \ref{fig:zoom_z40_angle1} and \ref{fig:zoom_z40_uy} we show additional high-resolution zooms of velocity fields to analyze local accuracy near critical urban structures such as street canyons and waterfronts.

\end{document}